\documentclass[10pt,journal,compsoc]{IEEEtran}
\ifCLASSOPTIONcompsoc
  \usepackage[nocompress]{cite}
\else
  \usepackage{cite}
\fi

\usepackage{times}
\usepackage{amssymb}
\usepackage{xspace}
\usepackage{amsmath}
\usepackage{enumitem}
\usepackage{graphicx}
\usepackage{caption}
\usepackage{subcaption}
\usepackage{textcomp}
\usepackage[usenames,dvipsnames]{color}
\usepackage[activate]{pdfcprot}
\usepackage{tabularx}
\usepackage{booktabs}
\usepackage{bm}
\newcolumntype{P}[1]{>{\raggedright\arraybackslash}m{#1}}%
\newcolumntype{C}[1]{>{\centering\arraybackslash}m{#1}}%
\newcolumntype{R}[1]{>{\raggedleft\arraybackslash}m{#1}}%
\usepackage{url}
\usepackage{changes}
\colorlet{Changes@Color}{red}

\newcolumntype{"}{@{\hskip\tabcolsep\vrule width 2pt\hskip\tabcolsep}}

\newcommand{\numberpictures}{213,659\xspace}

\newcommand{\numbermannualannotation}{37,667\xspace}
\newcommand{\datasetname}{MPIIGaze\xspace}
\newcommand{\datasetnameann}{MPIIGaze+\xspace}
\newcommand{\methodname}{GazeNet\xspace}
\newcommand{\methodnameann}{GazeNet+\xspace}

\begin{document}

\title{MPIIGaze: Real-World Dataset and Deep Appearance-Based Gaze Estimation}

\author{Xucong~Zhang,
        Yusuke~Sugano\IEEEauthorrefmark{1},
        Mario~Fritz,~\IEEEmembership{}%
        Andreas~Bulling~\IEEEmembership{}%
\IEEEcompsocitemizethanks{
\IEEEcompsocthanksitem X. Zhang, M. Fritz, and A. Bulling are with the Max Planck Institute for Informatics, Saarland Informatics Campus, Saarbr{\"u}cken, Germany.\protect\\
E-mail: \{xczhang, mfritz, bulling\}@mpi-inf.mpg.de
\IEEEcompsocthanksitem Y. Sugano is with the Graduate School of
Information Science and Technology, Osaka University, Japan.
E-mail: sugano@ist.osaka-u.ac.jp
\IEEEcompsocthanksitem \IEEEauthorrefmark{1}Work conducted while at the Max Planck Institute for Informatics}
}

\IEEEtitleabstractindextext{%
\begin{abstract}
Learning-based methods are believed to work well for unconstrained gaze estimation, i.e.\ gaze estimation from a monocular RGB camera without assumptions regarding user, environment, or camera.
However, current gaze datasets were collected under laboratory conditions and methods were not evaluated across multiple datasets.
Our work makes three contributions towards addressing these limitations.
First, we present the \datasetname dataset, which contains \numberpictures full face images and corresponding ground-truth gaze positions collected from 15 users during everyday laptop use over several months.
An experience sampling approach ensured continuous gaze and head poses and realistic variation in eye appearance and illumination.
To facilitate cross-dataset evaluations, \numbermannualannotation images were manually annotated with eye corners, mouth corners, and pupil centres.
Second, we present an extensive evaluation of state-of-the-art gaze estimation methods on three current datasets, including \datasetname.
We study key challenges including target gaze range, illumination conditions, and facial appearance variation.
We show that image resolution and the use of both eyes affect gaze estimation performance, while head pose and pupil centre information are less informative.
Finally, we propose \methodname, the first deep appearance-based gaze estimation method.
\methodname improves on the state of the art by 22\% (from a mean error of 13.9 degrees to 10.8 degrees) for the most challenging cross-dataset evaluation.
\end{abstract}

}

\maketitle

\IEEEdisplaynontitleabstractindextext

\IEEEpeerreviewmaketitle

\IEEEraisesectionheading{\section{Introduction}
\label{sec:introduction}}

\IEEEPARstart{G}{aze} estimation is well established as a research topic in computer vision because of its relevance for several applications, such as gaze-based human-computer interaction\cite{majaranta2014eye} or visual attention analysis~\cite{sugano16_uist,sattar15cvpr}.
Most recent learning-based methods leverage large amounts of both real and synthetic training data~\cite{odobez2013person,schneider2014manifold,suganolearning,wood2016learning} for person-independent gaze estimation.
They have thus brought us one step closer to the grand vision of \textit{unconstrained gaze estimation}:
3D gaze estimation in everyday environments and without any assumptions regarding users' facial appearance, geometric properties of the environment and camera, or image formation properties of the camera itself.
Unconstrained gaze estimation using monocular RGB cameras is particularly promising given the proliferation of such cameras in portable devices~\cite{wood14_etra} and public displays~\cite{zhang13_chi}.

While learning-based methods have demonstrated their potential for person-independent gaze estimation, methods have not been evaluated across different datasets to properly study their generalisation capabilities.
In addition, current datasets have been collected under controlled laboratory conditions that are characterised by limited variability in appearance and illumination and the assumption of accurate head pose estimates.
These limitations not only bear the risk of significant dataset bias -- an important problem also identified in other areas in computer vision, such as object recognition~\cite{torralba2011unbiased} or salient object detection~\cite{li2014secrets}.
They also impede further progress towards unconstrained gaze estimation, given that it currently remains unclear how state-of-the-art methods perform on real-world images and across multiple datasets.

\begin{figure}[t]
\center
\includegraphics[width=\columnwidth]{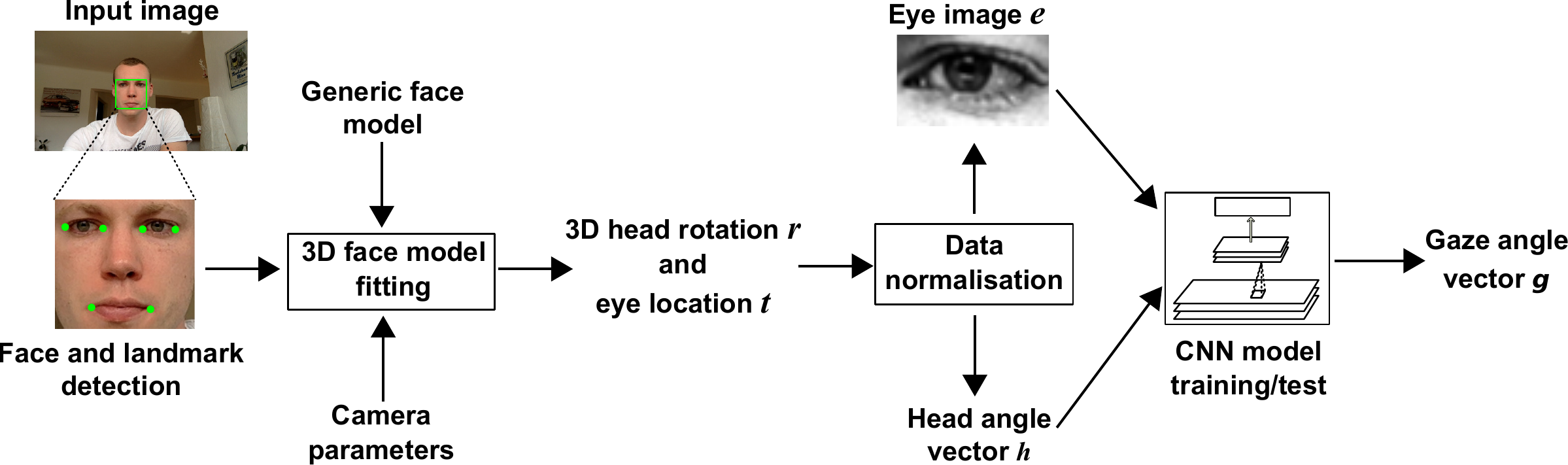}
\caption{Overview of \methodname -- appearance-based gaze estimation using a deep convolutional neural network (CNN).}
\label{fig:pipeline}
\end{figure}

\renewcommand{\tablename}{Fig.}
\setcounter{table}{1}
\setcounter{figure}{2}
\begin{table*}[ht]
\hskip-0.25cm\begin{tabularx}{\textwidth}{C{2.6cm} C{2.6cm} C{2.6cm} C{2.6cm} C{2.6cm} " C{2.6cm}}
\includegraphics[width=2.6cm]{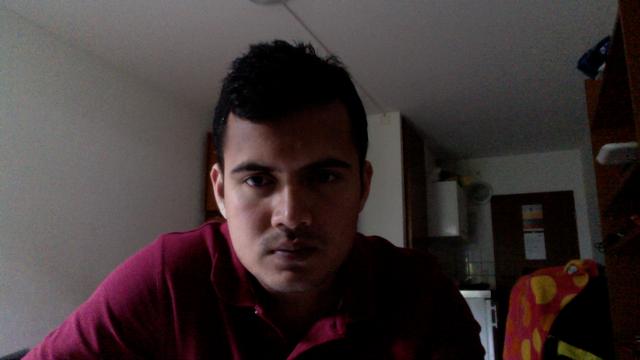} &
\includegraphics[width=2.6cm]{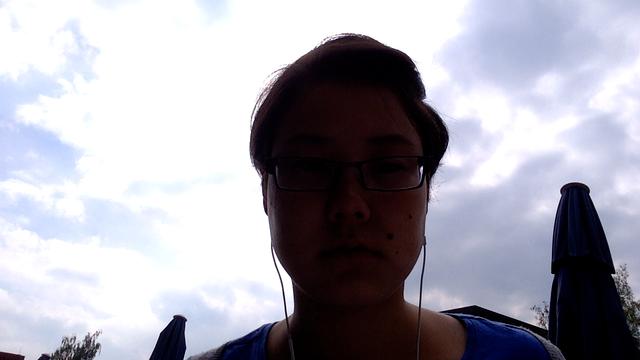} &
\includegraphics[width=2.6cm]{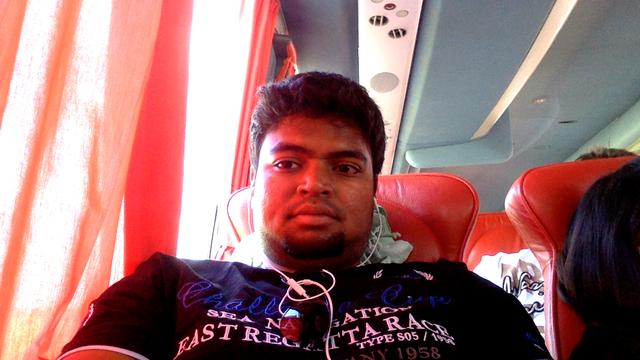} &
\includegraphics[width=2.6cm]{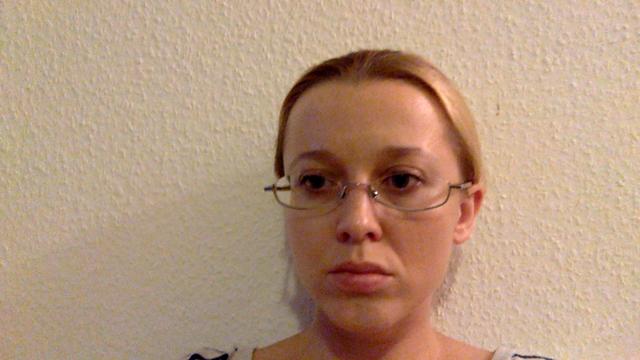} &
\includegraphics[width=2.6cm]{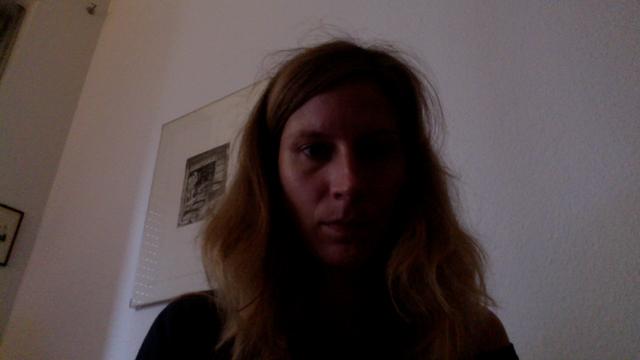} &
\includegraphics[width=2.6cm]{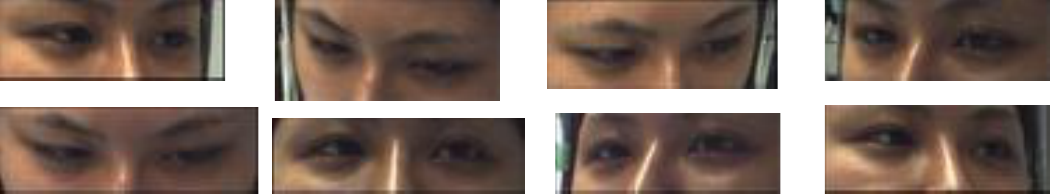} \\
\includegraphics[width=2.6cm]{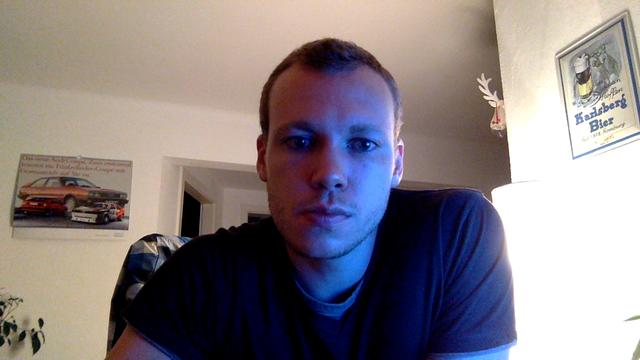} &
\includegraphics[width=2.6cm]{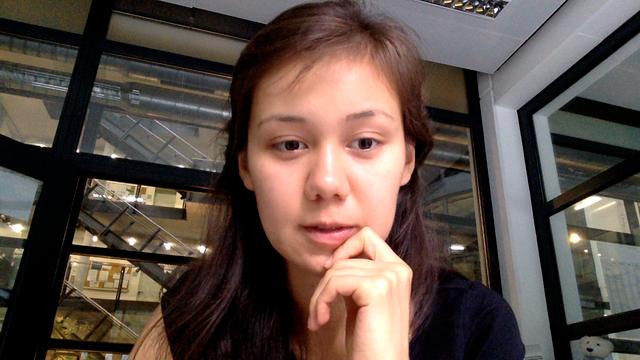} &
\includegraphics[width=2.6cm]{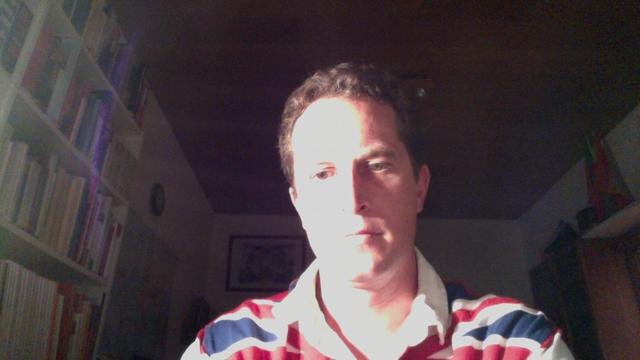} &
\includegraphics[width=2.6cm]{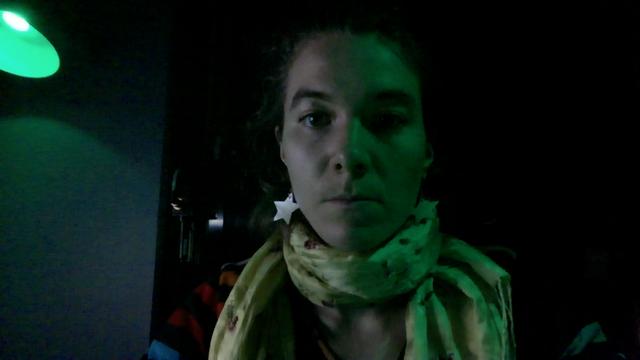} &
\includegraphics[width=2.6cm]{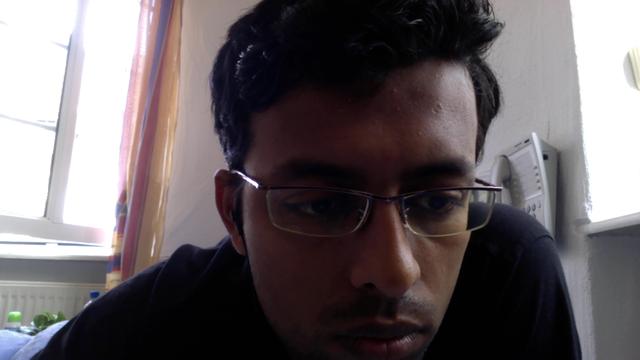} &
\includegraphics[height=1.5cm]{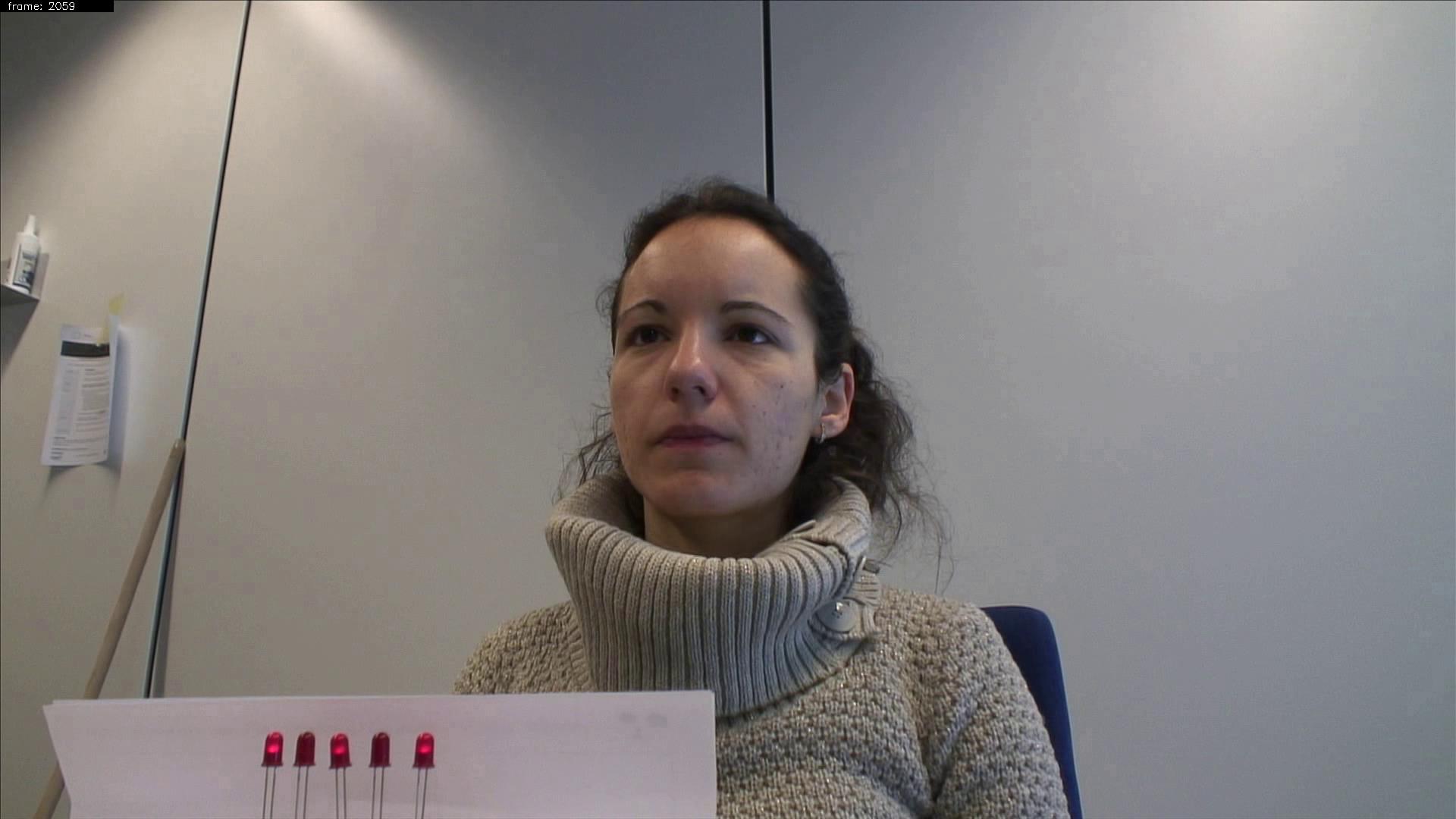} \\
\includegraphics[width=2.6cm]{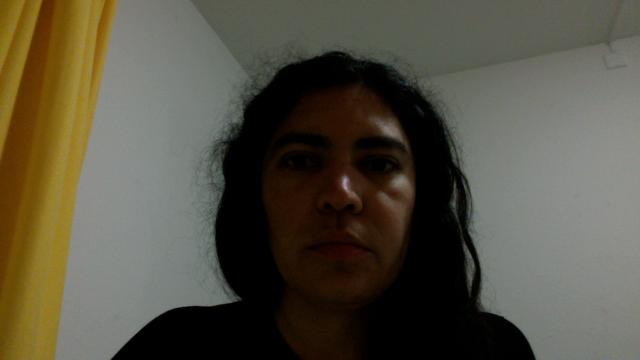} &
\includegraphics[width=2.6cm]{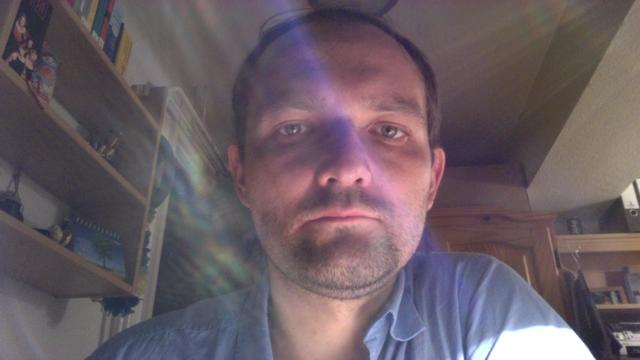} &
\includegraphics[width=2.6cm]{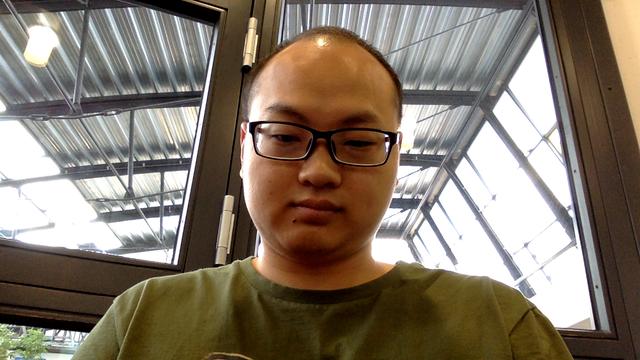} &
\includegraphics[width=2.6cm]{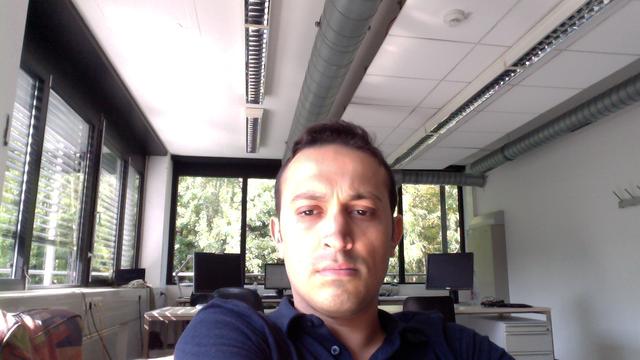} &
\includegraphics[width=2.6cm]{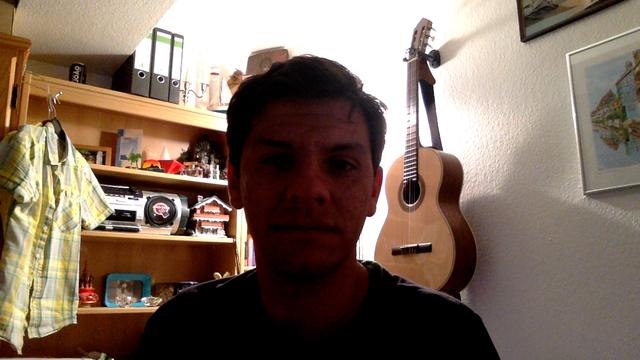} &
\includegraphics[height=1.5cm]{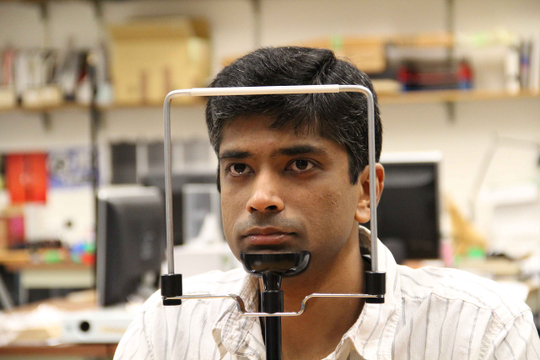} \\
\end{tabularx}
\caption{Sample images from the \datasetname dataset showing the considerable variability in terms of place and time of recording, eye appearance, and illumination (particularly directional light and shadows). For comparison, the last column shows sample images from other current publicly available datasets (cf. Table~\ref{tab:other_datasets}): UT Multiview~\cite{suganolearning} (top), EYEDIAP~\cite{FunesMora_ETRA_2014} (middle), and Columbia~\cite{smith2013gaze} (bottom).}
\label{fig:teaser}
\end{table*}
\renewcommand{\tablename}{Table}
\setcounter{table}{0}

This work aims to shed light on these questions and make the next step towards unconstrained gaze estimation.
To facilitate cross-dataset evaluations, we first introduce the \datasetname dataset, which contains \numberpictures images that we collected from 15 laptop users over several months in their daily life (see Fig.~\ref{fig:teaser}).
To ensure frequent sampling during this time period, we opted for an experience sampling approach in which participants were regularly triggered to look at random on-screen positions on their laptop.
This way, \datasetname not only offers an unprecedented realism in eye appearance and illumination variation but also in personal appearance -- properties not available in any existing dataset.
Methods for unconstrained gaze estimation have to handle significantly different 3D geometries between user, environment, and camera.
To study the importance of such geometry information, we ground-truth annotated \numbermannualannotation images with six facial landmarks (eye and mouth corners) and pupil centres.
These annotations make the dataset also interesting for closely related computer vision tasks, such as pupil detection.
The full dataset including annotations is available at \url{https://www.mpi-inf.mpg.de/MPIIGaze}.

Second, we conducted an extensive evaluation of several state-of-the-art methods on three current datasets: \datasetname, EYEDIAP~\cite{FunesMora_ETRA_2014}, and UT Multiview~\cite{suganolearning}.
We include a recent learning-by-synthesis approach that trains the model with synthetic data and fine-tunes it on real data~\cite{wood2015_iccv}.
We first demonstrate the significant performance gap between previous within- and cross-dataset evaluation conditions.
We then analyse various challenges associated with the unconstrained gaze estimation task, including gaze range, illumination conditions, and personal differences.
Our experiments show these three factors are responsible for 25\%, 35\% and 40\% performance gap respectively, when extending or restricting the coverage of training data.
These analyses reveal that, although largely neglected in previous research, illumination conditions represent an important source of error, comparable to differences in personal appearance.

Finally, we propose \methodname, the first deep appearance-based gaze estimation method based on a 16-layer VGG deep convolutional neural network.
\methodname outperforms the state of the art by 22\% on \datasetname and 8\% on EYEDIAP for the most difficult cross-dataset evaluation.
Our evaluations represent the first account of the state of the art in cross-dataset gaze estimation and, as such, provide valuable insights for future research on this important but so far under-investigated computer vision task.

An earlier version of this work was published in~\cite{zhang2015appearance}.
Parts of the text and figures are reused from that paper.
The specific changes implemented in this work are: 1) Extended annotation of 37,667 images with six facial landmarks (four eye corners and two mouth corners) and pupil centres, 2) updated network architecture to a 16-layer VGGNet, 3) new cross-dataset evaluation when training on synthetic data, 4) new evaluations of key challenges of the domain-independent gaze estimation task, specifically differences in gaze range, illumination conditions, and personal appearance, and 5) new evaluations on the influence of image resolution, the use of both eyes, and the use of head pose and pupil centre information on gaze estimation performance.

\section{Related Work}\label{sec:relatedwork}

\begin{table*}[ht]
\centering
\begin{tabularx}{\textwidth}{P{2.8cm} C{1.5cm} C{1.75cm} C{1.5cm} C{1.5cm} C{1.5cm} C{1.7cm} C{1.5cm} C{1.0cm}}
\toprule
& \textbf{Participants} & \textbf{Head poses} & \textbf{Gaze targets} & \textbf{Illumination conditions} & \textbf{Face annotations} & \textbf{Amount of data} & \textbf{Collection duration} & \textbf{3D anno.}\\
\midrule
Villaneuva et al.~\cite{villanueva2013hybrid} & 103 & 1 & 12 & 1 & 1,236 & 1,236 & 1 day & No\\
\midrule
TabletGaze~\cite{Huang2017} & 51 & continuous & 35 & 1 & none & 1,428 min & 1 day & No\\
\midrule
GazeCapture~\cite{krafka2016eye} & 1,474 & continuous & continuous & daily life & none & 2,445,504 & 1 day & No\\
\midrule
\midrule
Columbia~\cite{smith2013gaze} & 56 & 5 & 21 & 1 & none & 5,880& 1 day & Yes\\
\midrule
McMurrough et al.~\cite{McMurrough:2012:ETD:2168556.2168622} & 20 & 1 & 16 & 1 & none & 97 min & 1 day & Yes\\
\midrule
Weidenbacher et al.~\cite{weidenbacher07} & 20 & 19 & 2-9 & 1 & 2,220 & 2,220 & 1 day & Yes\\
\midrule
OMEG~\cite{he2015omeg} & 50 & 3 + continuous & 10 & 1 & unknown & 333 min & 1 day & Yes\\
\midrule
EYEDIAP~\cite{FunesMora_ETRA_2014} & 16 & continuous & continuous & 2 & none & 237 min & 2 days & Yes\\
\midrule
UT Multiview~\cite{suganolearning} & 50 & 8 + synthesised & 160 & 1 & 64,000 & 64,000 & 1 day & Yes\\
\midrule
\textbf{\datasetname (ours)} & \textbf{15} & \textbf{continuous} & \textbf{continuous} & \textbf{daily life} & \textbf{\numbermannualannotation} & \textbf{\numberpictures} & \textbf{9 days $\sim$ \newline 3 months} & \textbf{Yes}\\
\bottomrule
\end{tabularx}
\caption{Overview of publicly available appearance-based gaze estimation datasets showing the number of participants, head poses and on-screen gaze targets (discrete or continuous), illumination conditions, images with annotated face and facial landmarks, amount of data (number of images or duration of video), collection duration per participant, as well as the availability of 3D annotations of gaze directions and head poses. Datasets suitable for cross-dataset evaluation (i.e.\ that have 3D annotations) are listed below the double line.}
\label{tab:other_datasets}
\end{table*}

\subsection{Gaze Estimation Methods}
Gaze estimation methods can generally be distinguished as model-based or appearance-based~\cite{hansen2010eye}.
Model-based methods use a geometric eye model and can be further divided into corneal-reflection and shape-based methods. Corneal-reflection methods rely on eye features detected using reflections of an external infrared light source on the outermost layer of the eye, the cornea.
Early works on corneal reflection-based methods were limited to stationary settings~\cite{morimoto2002detecting,shih2004novel,yoo2005novel,hennessey2006single} but were later extended to handle arbitrary head poses using multiple light sources or cameras~\cite{zhu2005eye,zhu2006nonlinear}.
Shape-based methods~\cite{Ishikawa_2004_4705,Chen2008,yamazoe2008remote,valenti2012combining} infer gaze directions from the detected eye shape, such as the pupil centres or iris edges.
Although model-based methods have recently been applied to more practical application scenarios~\cite{jianfeng2014eye,mora2014geometric,sun2014realtime,wood14_etra,cristina2016model}, their gaze estimation accuracy is still lower, since they depend on accurate eye feature detections for which high-resolution images and homogeneous illumination are required.
These requirements have largely prevented these methods from being widely used in real-world settings or on commodity devices.

In contrast, appearance-based gaze estimation methods do not rely on feature point detection but directly regress from eye images to 3D gaze directions.
While early methods assumed a fixed head pose~\cite{baluja1994non,tan2002appearance,williams2006sparse,sewell10_chi,lu2014alr,Liang:2013:AGT:2509315.2509318}, more recent methods allow for free 3D head movement in front of the camera~\cite{lu2014learning,lu2012head,funes2012gaze,choi2013appearance,gao2014computer}.
Because they do not rely on any explicit shape extraction stage, appearance-based methods can handle low-resolution images and long distances.
However, these methods require more person-specific training data than model-based approaches to cover the significant variability in eye appearance caused by free head motion and were therefore mainly evaluated for a specific domain or person.
An open research challenge in gaze estimation is to learn gaze estimators that do not make any assumptions regarding the user, environment, or camera.

\subsection{Person-Independent Gaze Estimation}

The need to collect person-specific training data represents a fundamental limitation for both model-based and appearance-based methods.
To reduce the burden on the user, several previous works used interaction events, such as mouse clicks or key presses, as a proxy for users' on-screen gaze position~\cite{sugano2008incremental,Huang:2014:BSE:2647868.2655031}.
Alternatively, visual saliency maps~\cite{chen2011probabilistic,sugano2013appearance} or pre-recorded human gaze patterns of the presented visual stimuli~\cite{alnajarcalibration} were used as probabilistic training data to learn the gaze estimation function.
However, the need to acquire user input fundamentally limits the applicability of these approaches to interactive settings.

Other methods aimed to learn gaze estimators that generalise to arbitrary persons without requiring additional input.
A large body of works focused on cross-person evaluations in which the model is trained and tested on data from different groups of participants.
For example, Schneider et al.\ performed a cross-person evaluation on the Columbia dataset~\cite{smith2013gaze} with 21 gaze points for one frontal head pose of 56 participants~\cite{schneider2014manifold}.
Funes et al.\ followed a similar approach, but only evaluated on five participants~\cite{odobez2013person}.
To reduce data collection and annotation efforts, Sugano et al.\ presented a clustered random forest method that was trained on a large number of synthetic eye images~\cite{suganolearning}.
The images were synthesised from a smaller number of real images captured using a multi-camera setup and controlled lighting in a laboratory. 
Later works evaluated person-independent gaze estimation methods on the same dataset~\cite{yu2016learning,jeni2016person}.
Krafka et al.\ recently presented a method for person-independent gaze estimation that achieved 1.71 cm on an iPhone and 2.53 cm on an iPad screen~\cite{krafka2016eye}. However, the method assumed a fixed camera-screen relationship and therefore cannot be used for cross-dataset gaze estimation.

\subsection{Unconstrained Gaze Estimation}

Despite significant advances in person-independent gaze estimation, all previous works assumed training and test data to come from the same dataset.
We were first to study the practically more relevant but also significantly more challenging task of unconstrained gaze estimation via cross-dataset evaluation~\cite{zhang2015appearance}.
We introduced a method based on a multimodal deep convolutional neural network that outperformed all state-of-the-art methods by a large margin.
More recently, we proposed another method that, in contrast to a long-standing line of work in computer vision, only takes the full face image as input, resulting again in significant performance improvements for both 2D and 3D gaze estimation~\cite{zhang17_cvprw}.
In later works, Wood et al.\ demonstrated that large-scale methods for unconstrained gaze estimation could benefit from parallel advances in computer graphics techniques for eye region modelling.
These models were used to synthesise large amounts of highly realistic eye region images, thereby significantly reducing both data collection and annotation efforts~\cite{wood2015_iccv}.
Their latest model is fully morphable~\cite{wood16_eccv} and can synthesise large numbers of images in a few hours on commodity hardware~\cite{wood2016learning}.

\subsection{Datasets}

Several gaze estimation datasets have been published in recent years (see Table~\ref{tab:other_datasets} for an overview).
Early datasets were severely limited with respect to variability in head poses, on-screen gaze targets, illumination conditions, number of images, face and facial landmark annotations, collection duration per participant, and annotations of 3D gaze directions and head poses~\cite{McMurrough:2012:ETD:2168556.2168622,villanueva2013hybrid,weidenbacher07,smith2013gaze}.
\begin{figure*}[t]
        \centering
        \begin{subfigure}[b]{0.33\textwidth}
                \includegraphics[width=\textwidth]{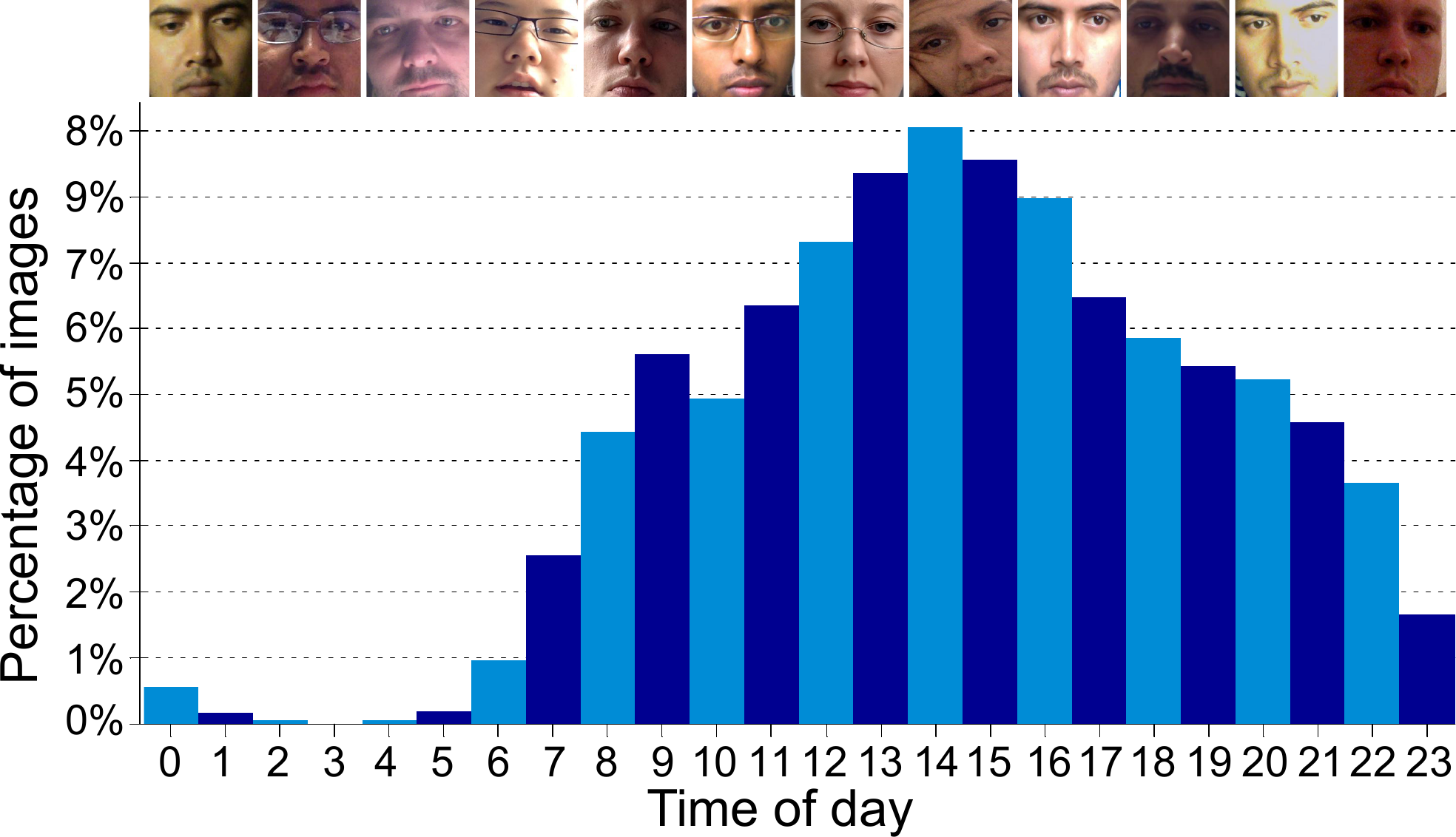}
                \label{fig:characteristics:timeofday}
        \end{subfigure}
        \begin{subfigure}[b]{0.33\textwidth}
                \includegraphics[width=\textwidth]{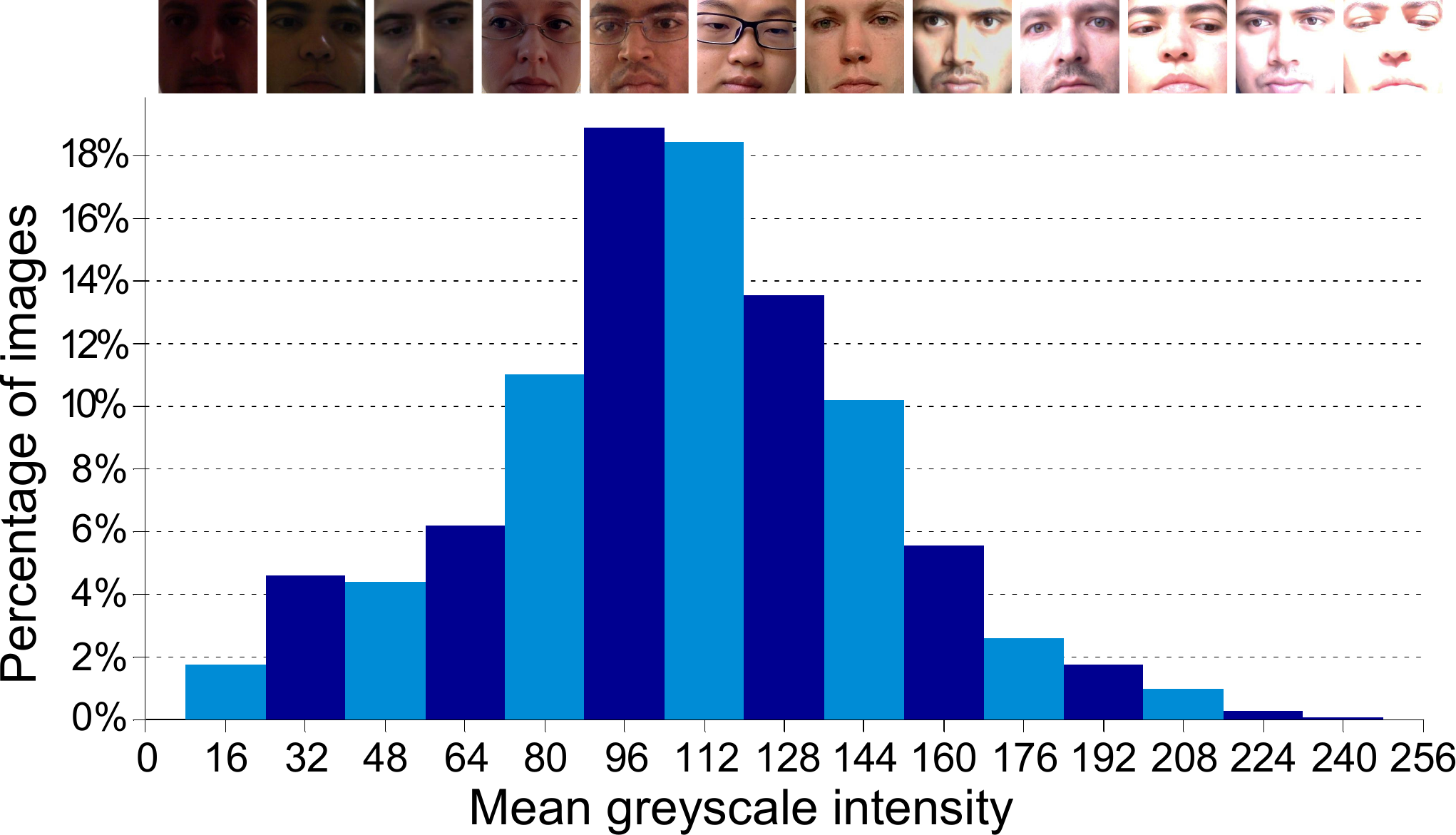}
                \label{fig:characteristics:intensity}
        \end{subfigure}
        \begin{subfigure}[b]{0.33\textwidth}
                \includegraphics[width=\textwidth]{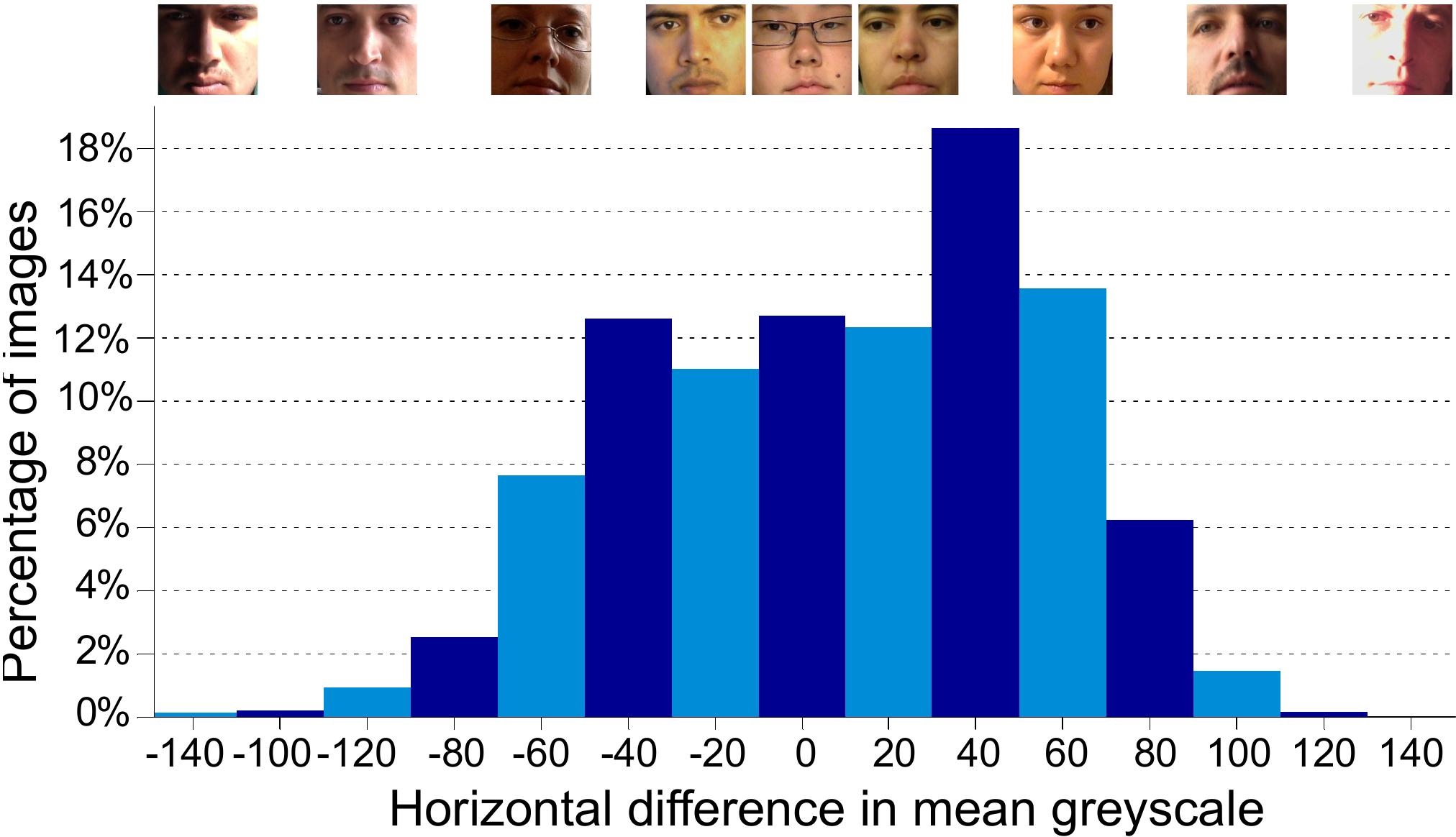}
                \label{fig:characteristics:directional}
        \end{subfigure}
        \caption{Key characteristics of our dataset. Percentage of images collected at different        times of day (left), having different mean grey-scale intensities within the face region (middle), and          having horizontally different mean grey-scale intensities between the left to right half of the face            region (right). Representative sample images are shown at the top.}
    \label{fig:characteristics}
\end{figure*}
More recent datasets are larger and cover the head pose and gaze ranges continuously.
The OMEG dataset includes 200 image sequences from 50 people with fixed and free head movement but discrete visual targets~\cite{he2015omeg}.
TabletGaze includes 16 videos recorded from 51 people looking at different points on a tablet screen~\cite{Huang2017}.
The EYEDIAP dataset contains 94 video sequences of 16 participants who looked at three different targets (discrete and continuous markers displayed on a monitor, and floating physical targets) under both static and free head motion conditions~\cite{FunesMora_ETRA_2014}.
The UT Multiview dataset also contains dense gaze samples of 50 participants and 3D reconstructions of eye regions that can be used to synthesise images for arbitrary head poses and gaze targets~\cite{suganolearning}.
However, all of these datasets were still recorded under controlled laboratory settings and therefore only include a few illumination conditions.
While the recent GazeCapture dataset~\cite{krafka2016eye} includes a large number of participants, the limited number of images and similar illumination conditions per participant make it less interesting for unconstrained gaze estimation. Even more importantly, the lack of 3D annotations limits its use to within-dataset evaluations.
Several large-scale datasets were published for visual saliency prediction, such as the crowd-sourced iSUN dataset~\cite{xu2015turkergaze}, but their focus is on bottom-up saliency prediction, and input face or eye images are not available.

\begin{figure}[t]
  \centering
        \begin{subfigure}[b]{0.33\linewidth}
                \includegraphics[width=\textwidth]{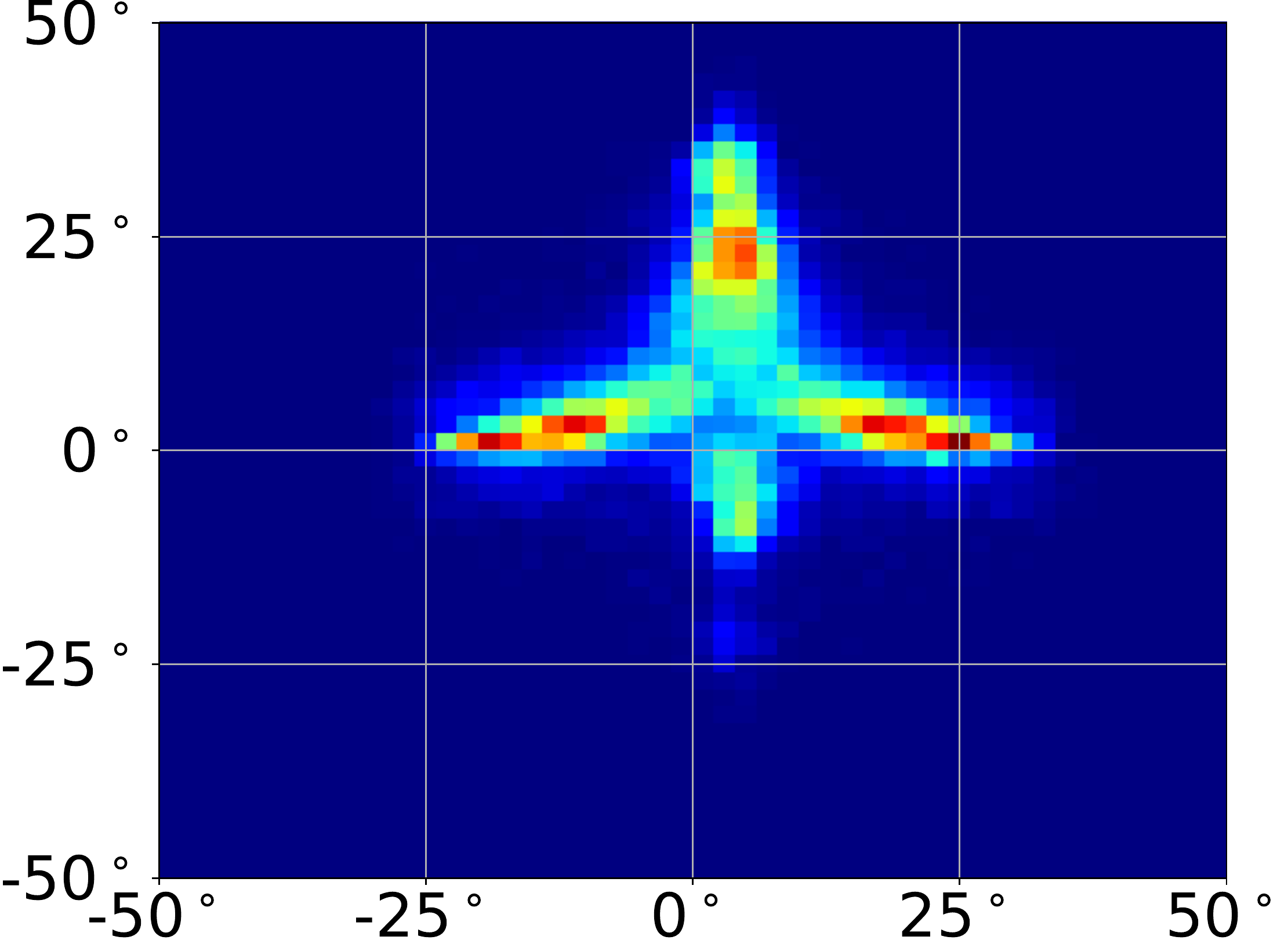}
                \caption{$\bm{h}$ (\datasetname)}
                \label{fig:dataset_comparison:mpii-pose}
        \end{subfigure}%
        \begin{subfigure}[b]{0.33\linewidth}
                \includegraphics[width=\textwidth]{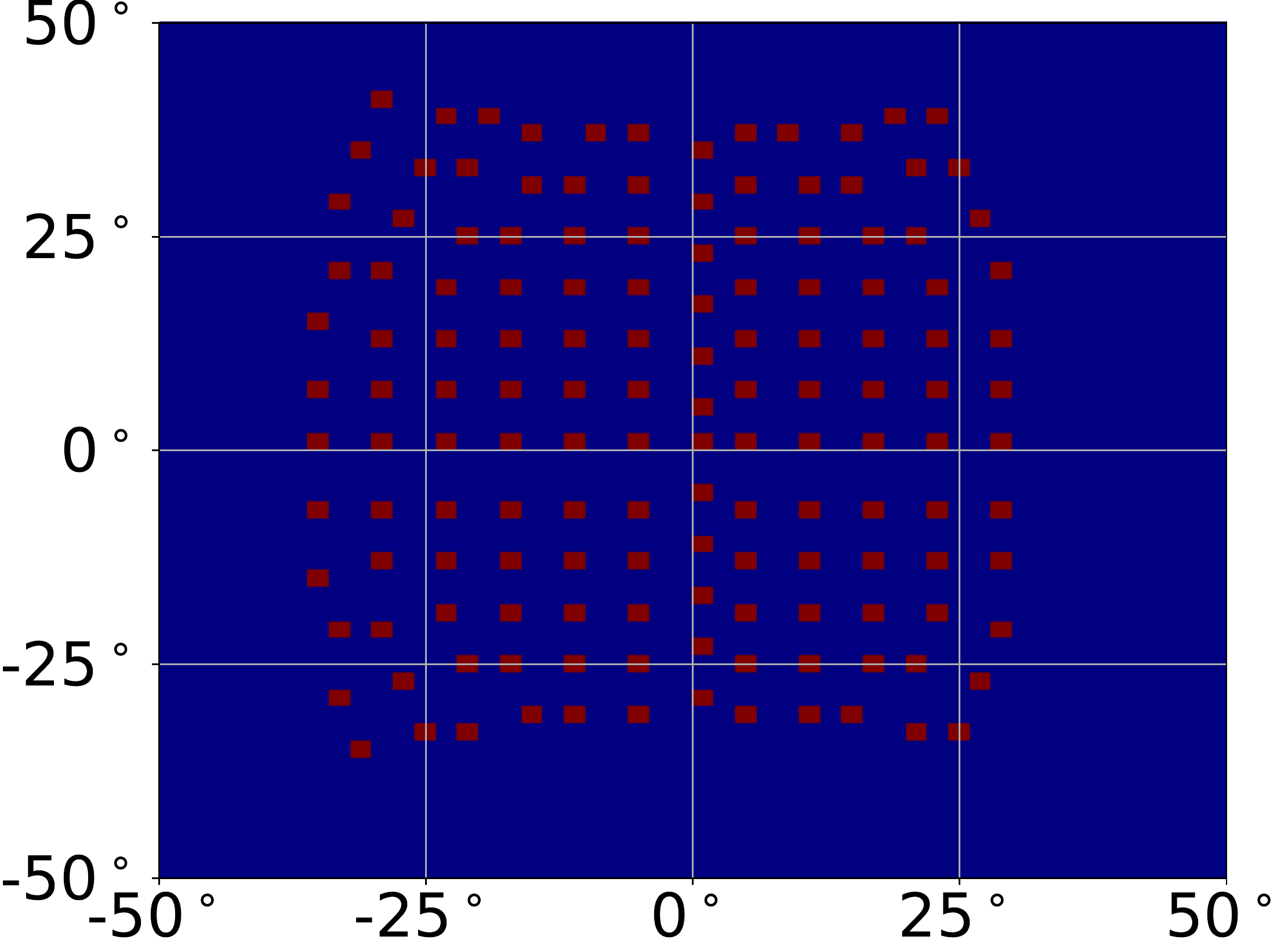}
                \caption{$\bm{h}$ (UT Multiview)}
                \label{fig:dataset_comparison:utfull-pose}
        \end{subfigure}%
        \begin{subfigure}[b]{0.33\linewidth}
                \includegraphics[width=\textwidth]{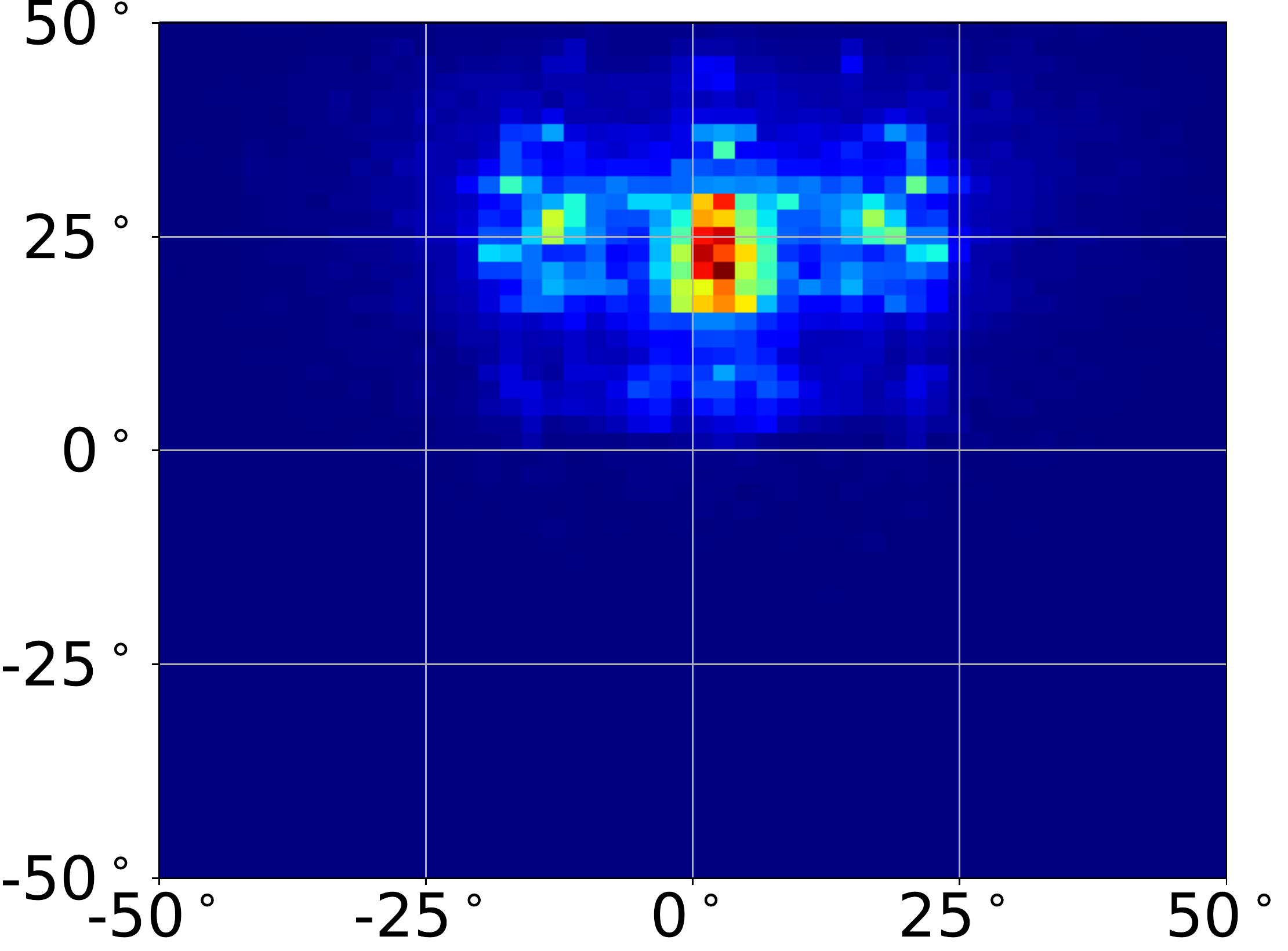}
                \caption{$\bm{h}$ (EYEDIAP)}
                \label{fig:dataset_comparison:idiap-ds-pose}
        \end{subfigure}%
        \vspace{1em}
        \begin{subfigure}[b]{0.33\linewidth}
                \includegraphics[width=\textwidth]{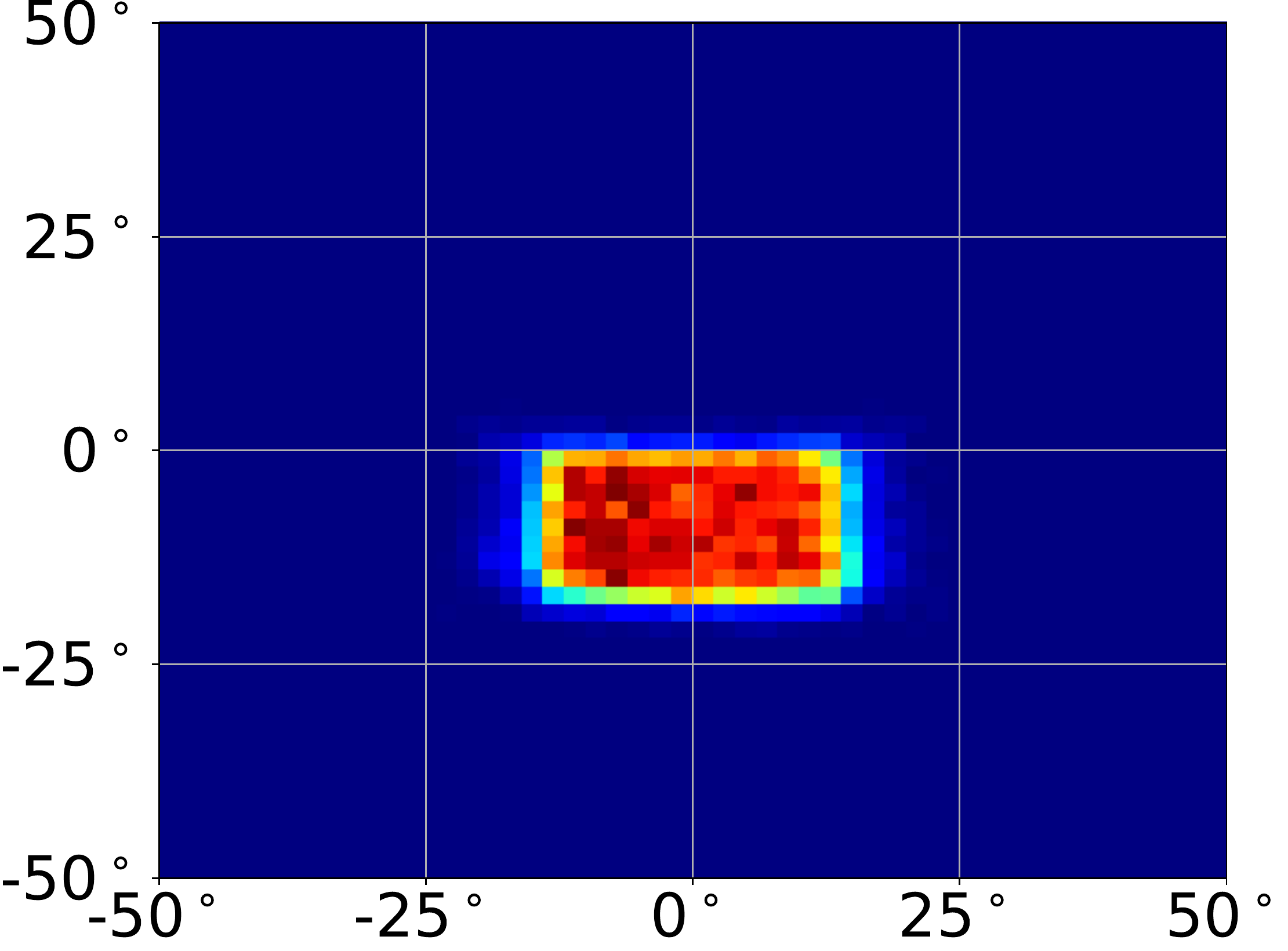}
                \caption{$\bm{g}$ (\datasetname)}
                \label{fig:dataset_comparison:mpii-gaze}
        \end{subfigure}%
        \begin{subfigure}[b]{0.33\linewidth}
                \includegraphics[width=\textwidth]{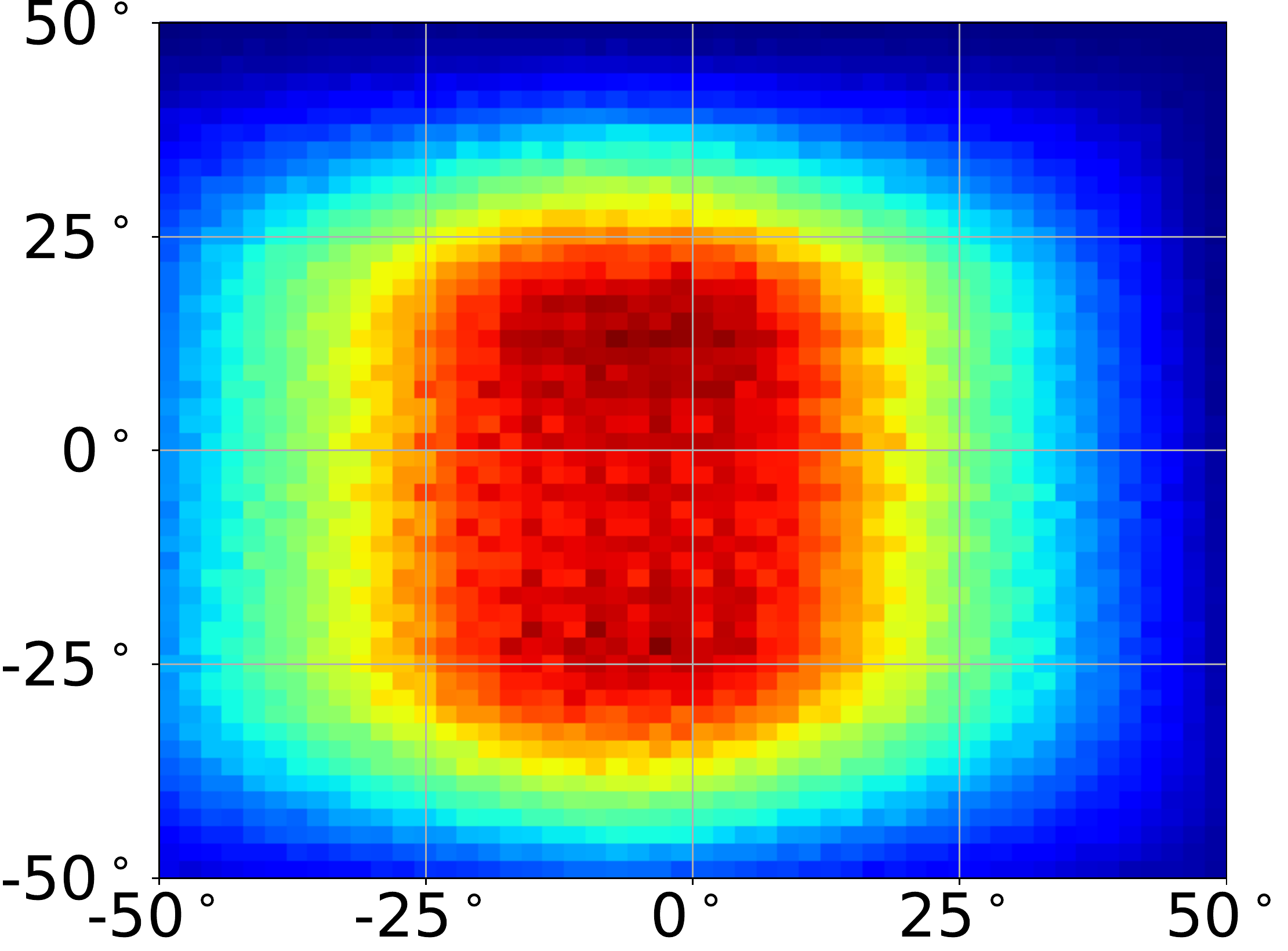}
                \caption{$\bm{g}$ (UT Multiview)}
                \label{fig:dataset_comparison:utfull-gaze}
        \end{subfigure}%
        \begin{subfigure}[b]{0.33\linewidth}
                \includegraphics[width=\textwidth]{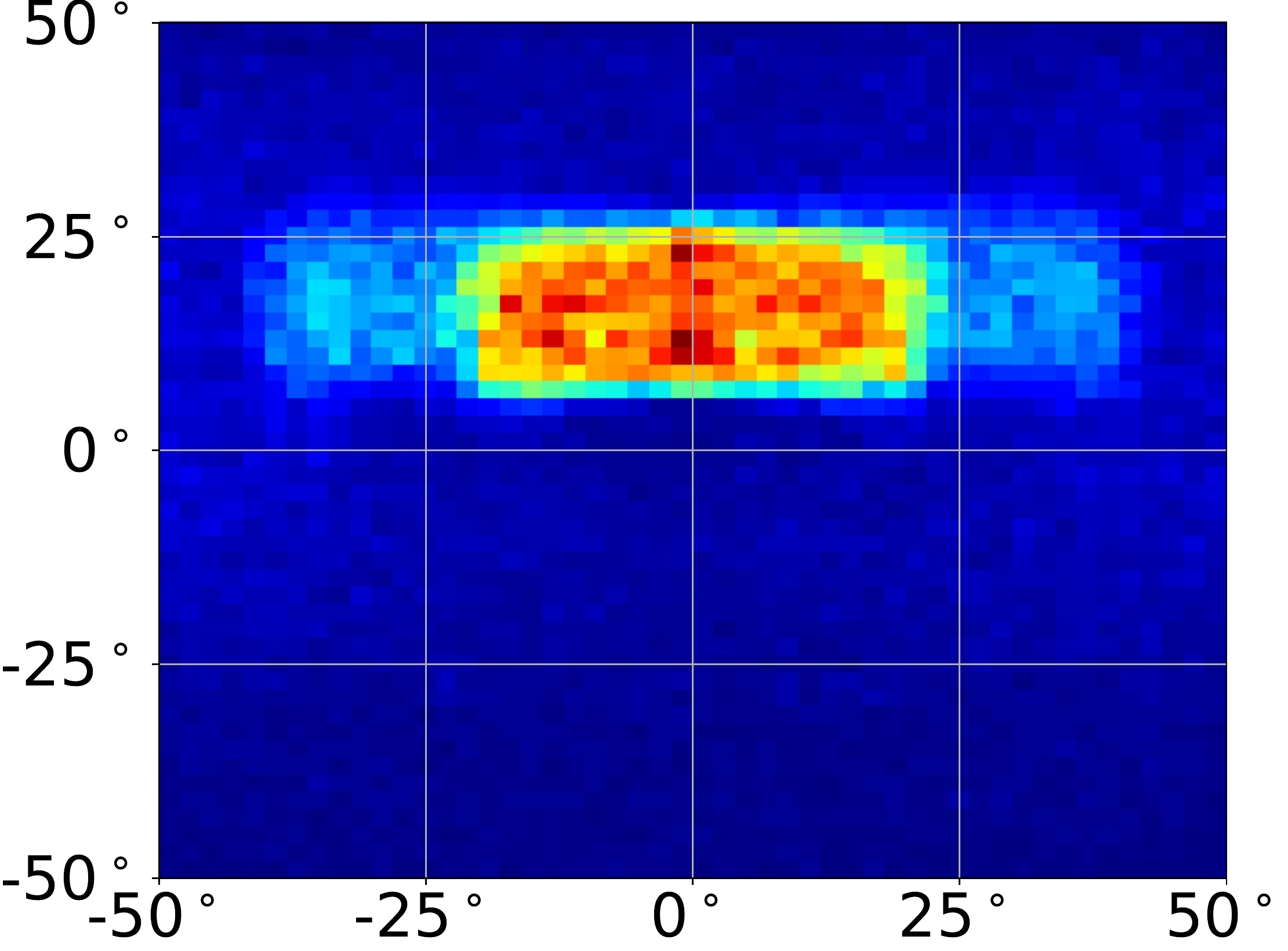}
                \caption{$\bm{g}$ (EYEDIAP)}
                \label{fig:dataset_comparison:idiap-ds-gaze}
        \end{subfigure}%
    \caption{Distributions of head angle ($\bm{h}$) and gaze angle ($\bm{g}$) in degrees for~\datasetname, UT Multiview, and the screen target sequences in EYEDIAP~ (cf. Table~\ref{tab:other_datasets}).}
    \label{fig:dataset_comparison}
\end{figure}
\section{The \datasetname~dataset}
\label{sec:Dataset}

To be able to evaluate methods for unconstrained gaze estimation, a dataset with varying illumination conditions, head poses, gaze directions, and personal appearance was needed.
To fill this gap, we collected the \datasetname dataset that contains a large number of images from different participants, covering several months of their daily life (see Fig.~\ref{fig:teaser} for sample images from our dataset).
The long-term recording resulted in a dataset that is one order of magnitude larger and significantly more variable than existing datasets (cf. Table~\ref{tab:other_datasets}).
All images in the dataset come with 3D annotations of gaze target and detected eye/head positions, which is required for cross-dataset training and evaluation.
Our dataset also provides manual facial landmark annotations on a subset of images, which enables a principled evaluation of gaze estimation performance and makes the dataset useful for other face-related tasks, such as eye or pupil detection.

\subsection{Collection Procedure}

We designed our data collection procedure with two main objectives in mind: 1) to record images of participants outside of controlled laboratory conditions, i.e during their daily routine, and 2) to record participants over several months to cover a wider range of recording locations and times, illuminations, and eye appearances. 
We opted for recording images on laptops not only because they are suited for long-term daily recordings but also because they are an important platform for eye tracking applications~\cite{majaranta2014eye}.
Laptops are personal devices, therefore typically remaining with a single user, and they are used throughout the day and over long periods of time. 
Although head pose and gaze range are a bit limited compared to the fully unconstrained case due to the screen size, they have a strong advantage in that the data recording can be carried out in a mobile setup.
They also come with high-resolution front-facing cameras and their large screen size allows us to cover a wide range of gaze directions.
We further opted to use an experience sampling approach to ensure images were collected regularly throughout the data collection period~\cite{larson1983experience}.

We implemented custom software running as a background service on participants' laptops, and opted to use the well-established moving dot stimulus\cite{kassner2014pupil}, to collect ground-truth annotations.
Every 10 minutes the software automatically asked participants to look at a random sequence of 20 on-screen positions (a recording session), visualised as a grey circle shrinking in size and with a white dot in the middle.
Participants were asked to fixate on these dots and confirm each by pressing the spacebar exactly once when the circle was about to disappear.
If they missed this small time window of about 500 ms, the software asked them to record the same on-screen location again right after the failure. 
While we cannot completely eliminate the possibility of bad ground truth, this approach ensured that participants had to concentrate and look carefully at each point during the recording.

Otherwise, participants were not constrained in any way, in particular as to how and where they should use their laptops.
Because our dataset covers different laptop models with varying screen size and resolution, on-screen gaze positions were converted to 3D positions in the camera coordinate system.
We obtained the intrinsic parameters from each camera beforehand using the camera calibration procedure from OpenCV~\cite{opencv_library}.
The 3D position of the screen plane in the camera coordinate system was estimated using a mirror-based calibration method in which the calibration pattern was shown on the screen and reflected to the camera using a mirror~\cite{Rodrigues2010Mirror}.
Both calibrations are required for evaluating gaze estimation methods across different devices.
3D positions of the six facial landmarks were recorded from all participants using an external stereo camera prior to the data collection, which could be used to build the 3D face model.

\subsection{Dataset Characteristics}

We collected a total of \numberpictures images from 15 participants (six female, five with glasses) aged between 21 and 35 years.
10 participants had brown, 4 green, and one grey iris colour.
Participants collected the data over different time periods ranging from 9 days to 3 months.
The number of images collected for each participant varied from 1,498 to 34,745.
Note that we only included images in which a face could be detected (see Section 4.1).
Fig.~\ref{fig:characteristics} (left) shows a histogram of times of the recording sessions.
Although there is a natural bias towards working hours, the figure shows the high variation in recording times.
Consequently, our dataset also covers significant variation in illumination.
To visualise the different illumination conditions, Fig.~\ref{fig:characteristics} (bottom) shows a histogram of mean grey-scale intensities inside the face region.
Fig.~\ref{fig:characteristics} (right) further shows a histogram of the mean intensity differences from the right side to the left side of the face region, indicative of strong directional light for a substantial number of images.

The 2D histograms in Fig.~\ref{fig:dataset_comparison} visualise the distributions of head and gaze angles $\bm{h}, \bm{g}$ in the normalised space, colour-coded from blue (minimum) to red (maximum), for \datasetname in comparison with two other recent datasets, EYEDIAP (all screen target sequences)~\cite{FunesMora_ETRA_2014} and UT Multiview~\cite{suganolearning} (see Section~\ref{subsec:normalisation} for a description of the normalisation procedure).
The UT Multiview dataset (see Fig.~\ref{fig:dataset_comparison:utfull-pose} and~\ref{fig:dataset_comparison:utfull-gaze}) was only recorded under a single controlled lighting condition, but provides good coverage of the gaze and pose spaces.
For the EYEDIAP dataset, Fig.~\ref{fig:dataset_comparison:idiap-ds-pose} and~\ref{fig:dataset_comparison:idiap-ds-gaze} show distributions of 2D screen targets that are comparable to our setting, yet gaze angle distributions do not overlap, due to different camera and gaze target plane setups (see Fig.~\ref{fig:dataset_comparison:mpii-pose} and~\ref{fig:dataset_comparison:mpii-gaze}).
For our \datasetname dataset, gaze directions tend to be below the horizontal axis in the camera coordinate system because the laptop-integrated cameras were positioned above the screen, and the recording setup biased the head pose to a near-frontal pose.
The gaze angles in our dataset are in the range of [-1.5, 20] degrees in the vertical and [-18, +18] degrees in the horizontal direction.

\begin{figure}[t]
  \centering
        \begin{subfigure}[b]{0.5\linewidth}
                \centering
                \includegraphics[width=0.9\textwidth]{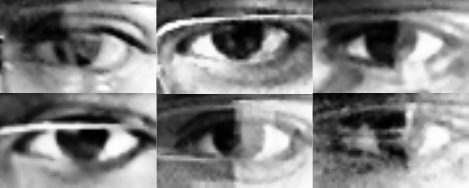}
                \caption{MPIIGaze gazing (0$^{\circ}$, 0$^{\circ}$)}
                \label{fig:dataset_comparison:mpii-images-glasses}
        \end{subfigure}%
        \begin{subfigure}[b]{0.5\linewidth}
                \centering
                \includegraphics[width=0.9\textwidth]{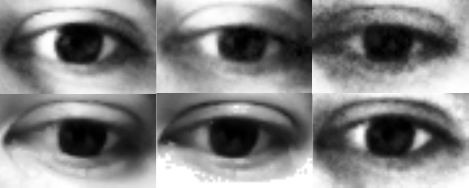}
                \caption{MPIIGaze gazing (15$^{\circ}$, -15$^{\circ}$)}
                \label{fig:dataset_comparison:mpii-images}
        \end{subfigure}%
        \vspace{1em}
        \begin{subfigure}[b]{0.5\linewidth}
                \centering
                \includegraphics[width=0.9\textwidth]{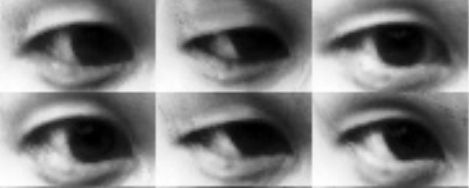}
                \caption{UT Multiview gazing (30$^{\circ}$, 5$^{\circ}$)}
                \label{fig:dataset_comparison:ut-images}
        \end{subfigure}%
        \begin{subfigure}[b]{0.5\linewidth}
                \centering
                \includegraphics[width=0.9\textwidth]{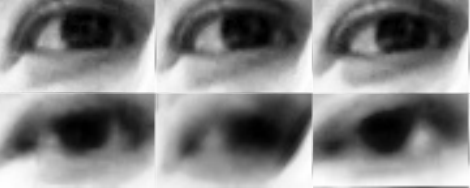}
                \caption{EYEDIAP gazing (25$^{\circ}$, 15$^{\circ}$)}
                \label{fig:dataset_comparison:idiap-images}
        \end{subfigure}%

    \caption{Sample images from a single person for roughly the same gaze directions from \datasetname with (a) and without (b) glasses, UT Multiview (c), and EYEDIAP (d).}
    \label{fig:dataset_comparison_image}
\end{figure}

Finally, Fig.~\ref{fig:dataset_comparison_image} shows sample eye images from each dataset after normalisation.
Each group of images was randomly selected from a single person for roughly the same gaze directions.
Compared to the UT Multiview and EYEDIAP datasets (see Fig.~\ref{fig:dataset_comparison:ut-images} and~\ref{fig:dataset_comparison:idiap-images}), \datasetname contains larger appearance variations even inside the eye region (see Fig.~\ref{fig:dataset_comparison:mpii-images}), particularly for participants wearing glasses (see Fig.~\ref{fig:dataset_comparison:mpii-images-glasses}).

\subsection{Facial Landmark Annotation}

We manually annotated a subset of images with facial landmarks to be able to evaluate the impact of face alignment errors on gaze estimation performance.
To this end, we annotated the evaluation subset used in~\cite{zhang2015appearance} that consists of a randomly-selected 1,500 left eye and 1,500 right eye images of all 15 participants.
Because eye images could be selected from the same face, this subset contains a total of 37,667 face images.

\begin{figure}[t]
\begin{subfigure}[b]{0.25\linewidth}
    \centering
    \includegraphics[width=\textwidth]{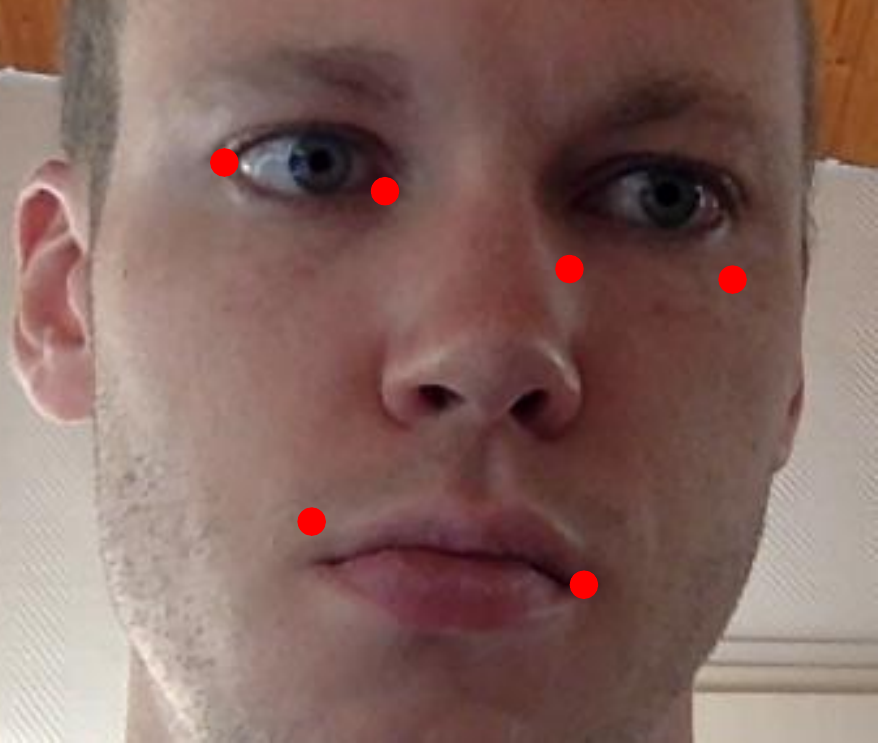}
    \caption{}
    \label{fig:detected_face}
\end{subfigure}
\begin{subfigure}[b]{0.25\linewidth}
    \centering
    \includegraphics[width=\textwidth]{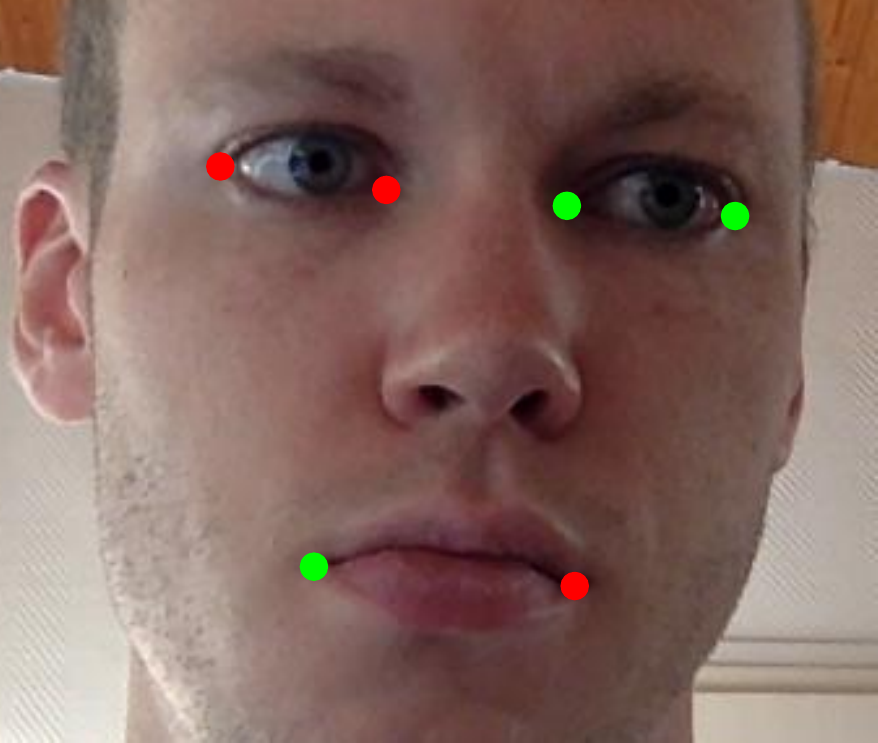}
    \caption{}
    \label{fig:annotation_face}
\end{subfigure}
\begin{subfigure}[b]{0.45\linewidth}
    \centering
    \includegraphics[width=\textwidth]{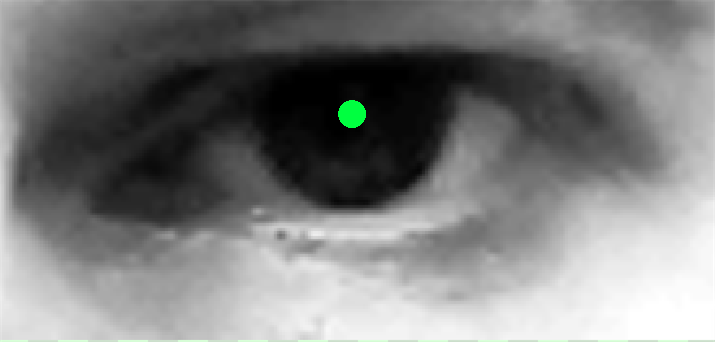}
    \caption{}
    \label{fig:annotation_pupil}
\end{subfigure}
\caption{We manually annotated \numbermannualannotation images with seven facial landmarks: the corners of the left and right eye, the mouth corners, and the pupil centres. We used a semi-automatic annotation approach: (a) Landmarks were first detected automatically (in red) and, (b) if needed, corrected manually post-hoc (in green). We also manually annotated the pupil centre without any detection (c). Note that this is only for completeness and we do not use the pupil centre as input for our method later.}
\label{fig:annotation}
\end{figure}

The annotation was conducted in a semi-automatic manner.
We first applied a state-of-the-art facial landmark detection method~\cite{Baltrusaitis2014CCNF}, yielding six facial landmarks per face image: the four eye and two mouth corners.
We then showed these landmarks to two experienced human annotators and asked them to flag those images that contained incorrect landmark locations or wrong face detections (see Fig.~\ref{fig:annotation_face}).
5,630 out of 37,667 images were flagged for manual annotation in this process.
Subsequently, landmark locations for all of these images were manually corrected by the same annotators.
Since automatic pupil centre localisation remains challenging~\cite{tonsen2016labelled}, we cropped the eye images using the manually-annotated facial landmarks and asked the annotators to annotate the pupil centres (see Fig.~\ref{fig:annotation_pupil}).

\begin{figure}[t]
\center
\includegraphics[width=0.9\columnwidth]{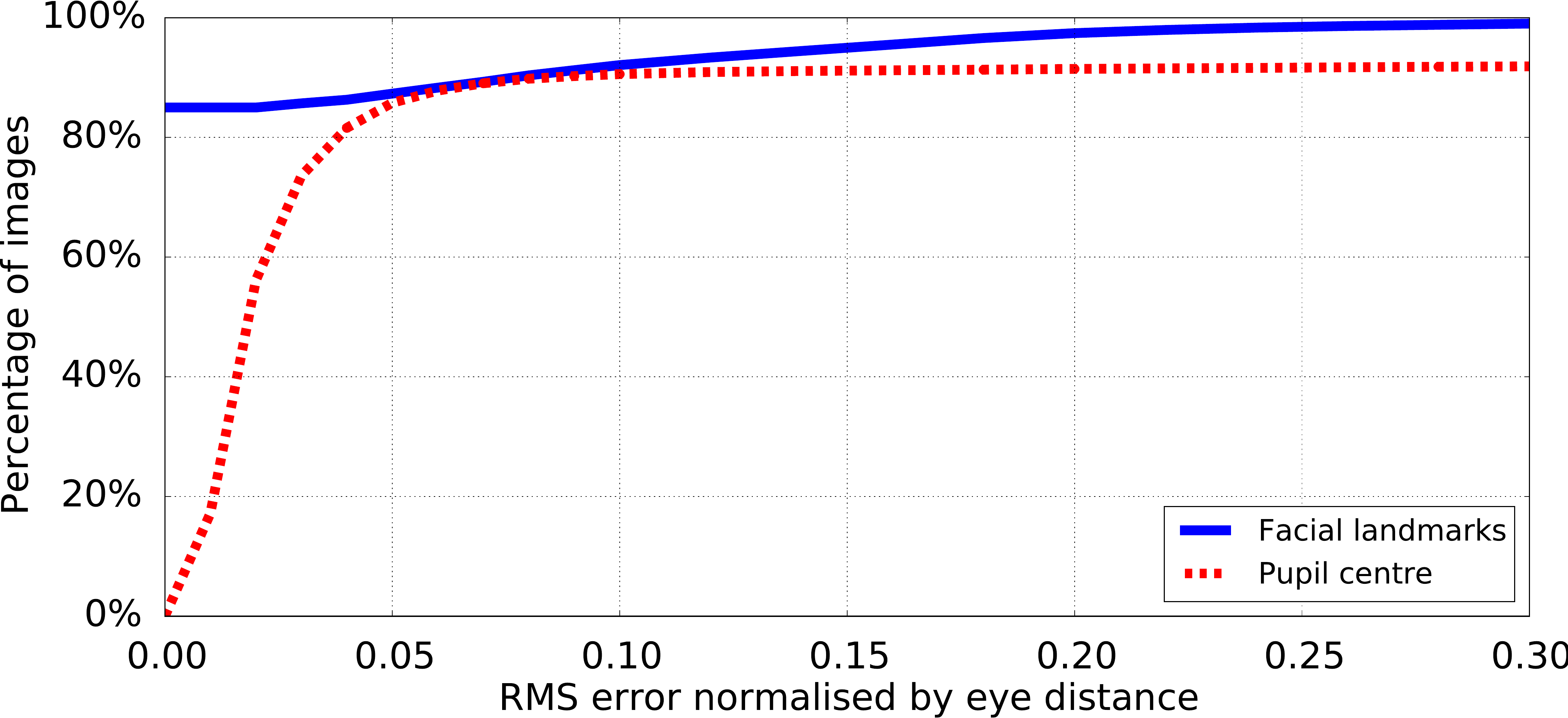}
\caption{Percentage of images for different error levels in the detection of facial landmarks (blue solid line) and pupil centres (red dashed line).
The x-axis shows the root-mean-square (RMS) distance between the detected and annotated landmarks, normalised by the distance between both eyes.
}
\label{fig:landmark_error}
\end{figure}

Fig.~\ref{fig:landmark_error} shows the detection error for facial landmarks and pupil centres when compared to the manual annotation. 
We calculated the error as the average root-mean-square (RMS) distances between the detected and annotated landmarks per face image.
As can be seen from the figure, 85\% of the images had no error in the detected facial landmarks.
0.98\% of the images had normalised RMS error less than 0.3.
This error roughly corresponds to the size of one eye and indicates that in these cases the face detection method failed to correctly detect the target face.
For the pupil centre (red line), the error for each eye image is the RMS between the detected and annotated pupil centre normalised by the distance between both eyes.
A normalised RMS error of 0.01 roughly corresponds to the size of the pupil, and 80\% of the images had lower pupil detection performance.

\section{Method}\label{sec:method}

Prior work performed person-independent gaze estimation using 2D regression in the screen coordinate system~\cite{Huang2017,krafka2016eye}.
Because this requires a fixed position of the camera relative to the screen, these methods are limited to the specific device configuration, i.e.\ do not directly generalise to other devices.
The recent success of deep learning combined with the availability of large-scale datasets, such as \datasetname, opens up promising new directions towards unconstrained gaze estimation that was not previously possible.
In particular, large-scale methods promise to learn gaze estimators that can handle the significant variability in domain properties as well as user appearance.
Fig.~\ref{fig:pipeline} shows an overview of our GazeNet method based on a multimodal convolutional neural network (CNN). 
We first use state-of-the-art face detection~\cite{dlib09} and facial landmark detection~\cite{Baltrusaitis2014CCNF} methods to locate landmarks in the input image obtained from the calibrated monocular RGB camera.
We then fit a generic 3D facial shape model to estimate 3D poses of the detected faces and apply the space normalisation technique proposed in~\cite{suganolearning} to crop and warp the head pose and eye images to the normalised training space.
A CNN is finally used to learn a mapping from the head poses and eye images to 3D gaze directions in the camera coordinate system.

\subsection{Face Alignment and 3D Head Pose Estimation}

\begin{figure}[t]
\center
\includegraphics[width=0.8\columnwidth]{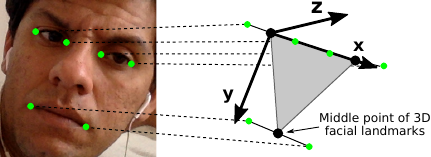}
\caption{Definition of the head coordinate system defined based on the triangle connecting three midpoints of the eyes and mouth. The x-axis goes through the midpoints of both while the y-axis is perpendicular to the x-axis inside the triangle plane. The z-axis is perpendicular to this triangle plane.}
\label{fig:head_pose_coordinate}
\end{figure}

Our method first detects the user's face in the image with a HOG-based method~\cite{dlib09}.
We assume a single face in the images and take the largest bounding box if the detector returned multiple face proposals.
We discard all images in which the detector fails to find any face, which happened in about 5\% of all cases.
Afterwards, we use a continuous conditional neural fields (CCNF) model framework to detect facial landmarks~\cite{Baltrusaitis2014CCNF}.

While previous works assumed accurate head poses, we use a generic mean facial shape model $\bm{F}$ for the 3D pose estimation to evaluate the whole gaze estimation pipeline in a practical setting.
The generic mean facial shape $\bm{F}$ is built as the averaged shape across all the participants, which could also be derived from any other 3D face models.
We use the same definition of the face model and head coordinate system as~\cite{suganolearning}.
The face model $\bm{F}$ consists of 3D positions of six facial landmarks (eye and mouth corners, cf. Fig.\ref{fig:pipeline}).
As shown in Fig.\ref{fig:head_pose_coordinate}, the right-handed head coordinate system is defined according to the triangle connecting three midpoints of the eyes and mouth. The x-axis is defined as the line connecting midpoints of the two eyes in the direction from the right eye to the left eye, and the y-axis is defined to be perpendicular to the x-axis inside the triangle plane in the direction from the eye to the mouth. The z-axis is hence perpendicular to the triangle, and pointing backwards from the face.
Obtaining the 3D rotation matrix $\bm{R}_r$ and translation vector $\bm{t}_r$ of the face model from the detected 2D facial landmarks $\bm{p}$ is a classical \textit{Perspective-n-Point}, problem which is estimating the 3D pose of an object given its 3D model and the corresponding 2D projections in the image.
We fit $\bm{F}$ to detected facial landmarks by estimating the initial solution using the EPnP algorithm~\cite{Lepetit2009EPnP} and further refine the pose by minimising the Levenberg-Marquardt distance.

\subsection{Eye Image Normalisation}
\label{subsec:normalisation}

\begin{figure}[t]
\center
\includegraphics[width=\columnwidth]{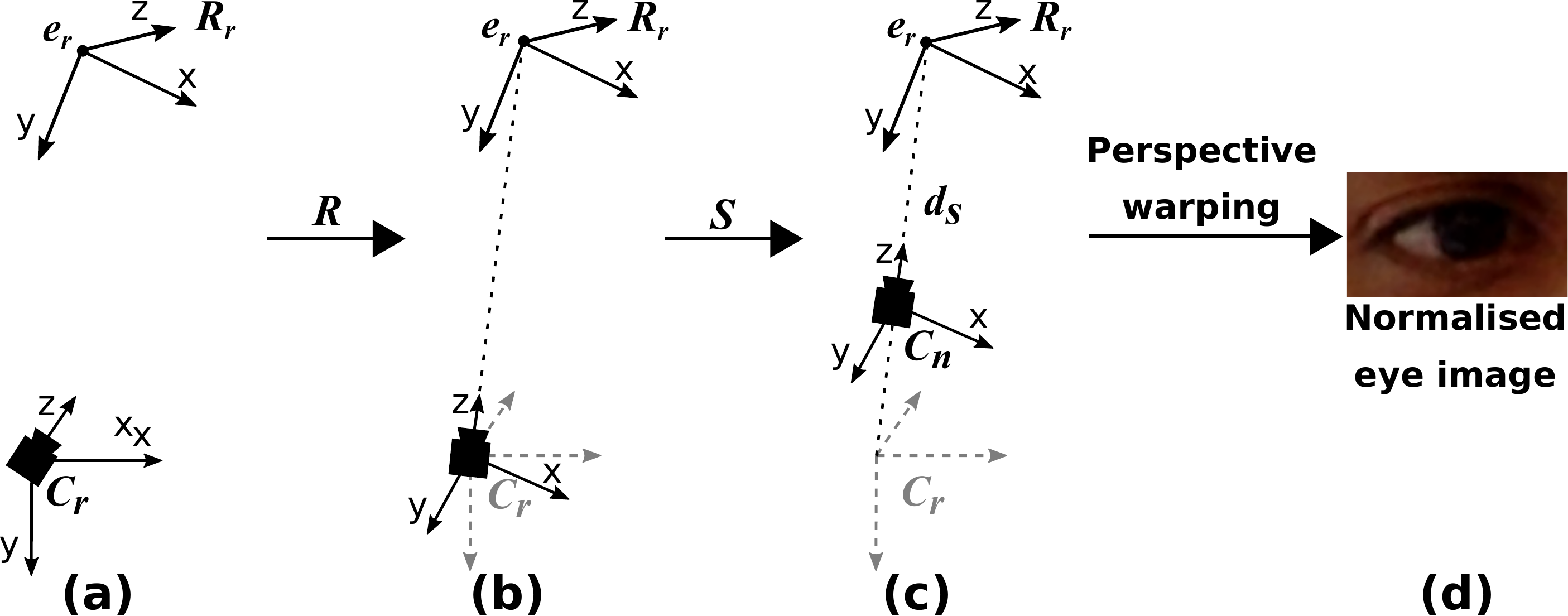}
\caption{Procedure for eye image normalisation. (a) Starting from the head pose coordinate system centred at one of the eye centres $\bm{e}_r$ (top) and the camera coordinate system (bottom); (b) the camera coordinate system is rotated with $\bm{R}$; (c) the head pose coordinate system is scaled with matrix $\bm{S}$; (d) the normalised eye image is cropped from the input image by the image transformation matrix corresponding to these rotations and scaling. }
\label{fig:data_normalisation}
\end{figure}

Given that our key interest is in cross-dataset evaluation, we normalise the image and head pose space as introduced in~\cite{suganolearning}.
Fundamentally speaking, object pose has six degrees of freedom, and in the simplest case the gaze estimator has to handle eye appearance changes in this 6D space.
However, if we assume that the eye region is planar, arbitrary scaling and rotation of the camera can be compensated for by its corresponding perspective image warping. Therefore, the appearance variation that needs to be handled inside the appearance-based estimation function has only two degrees of freedom.
The task of pose-independent appearance-based gaze estimation is to learn the mapping between gaze directions and eye appearances, which cannot be compensated for by virtually rotating and scaling the camera.

The detailed procedure for the eye image normalisation is shown in Fig.\ref{fig:data_normalisation}.
Given the head rotation matrix $\bm{R}_r$ and the eye position in the camera coordinate system $\bm{e}_r = \bm{t}_r + \bm{e}_h$ where $\bm{e}_h$ is the position of the midpoint of the two eye corners defined in the head coordinate system (Fig.~\ref{fig:data_normalisation} (a)), we need to compute the conversion matrix $\bm{M} = \bm{S}\bm{R}$ for normalisation.
As illustrated in Fig.~\ref{fig:data_normalisation} (b), $\bm{R}$ is the inverse of the rotation matrix that rotates the camera so that the the camera looks at $\bm{e}_r$ (i.e., the eye position is located along the $z$-axis of the rotated camera), the $x$-axis of the head coordinate system is perpendicular to the $y$-axis of the camera coordinate system.
The scaling matrix $\bm{S}=\mathrm{diag}(1,1,d_n/\lVert \bm{e}_r \rVert)$ (Fig.~\ref{fig:data_normalisation} (c)) is then defined so that the eye position $\bm{e}_r$ is located at a distance $d_n$ from the origin of the scaled camera coordinate system.

$\bm{M}$ describes a 3D scaling and rotation that brings the eye centre to a fixed position in the (normalised) camera coordinate system, and is used for interconversion of 3D positions between the original and the normalised camera coordinate system.
If we denote the original camera projection matrix obtained from camera calibration as $\bm{C}_r$ and the normalised camera projection matrix as $\bm{C}_n$, the same conversion can be applied to the original image pixels via perspective warping using the image transformation matrix $\bm{W} = \bm{C}_n\bm{M}\bm{C}^{-1}_r$ (Fig.~\ref{fig:data_normalisation} (d)).
$\bm{C}_n=[f_x, 0, c_x; 0, f_y, c_y; 0, 0, 1]$, 
where $f$ and $c$ indicate the focal length and principal point of the normalised camera, which are arbitrary parameters of the normalised space.
The whole normalisation process is applied to both right and left eyes in the same manner, with $\bm{e}_r$ defined according to the corresponding eye position.

This yields a set of an eye image $\bm{I}$, a head rotation matrix $\bm{R}_n = \bm{M}\bm{R}_r$, and a gaze angle vector $\bm{g}_n= \bm{M}\bm{g}_r$ in the normalised space.
$\bm{g}_r$ is the 3D gaze vector originating from $\bm{e}_r$ in the original camera coordinate system.
The normalised head rotation matrix $\bm{R}_n$ is then converted to a three-dimensional rotation angle vector $\bm{h}_n$.
Since rotation around the z-axis is always zero after normalisation, $\bm{h}_n$ can be represented as a two-dimensional rotation vector (horizontal and vertical orientations) $\bm{h}$.
$\bm{g}_n$ is also represented as a two-dimensional rotation vector $\bm{g}$ assuming a unit length.
We define $d_n$ to be 600 mm and focal length $f_x$ and $f_y$ of the normalised camera projection matrix $\bm{C}_n$ to be 960, so that it is compatible with the UT Multiview dataset~\cite{suganolearning}.
The resolution of the normalised eye image is set to $\bm{I}$ in $60 \times 36$ pixels, and thus $c_x$ and $c_y$ are set to 30 and 18, respectively.
Eye images $\bm{I}$ are converted to grey scale and histogram-equalised after normalisation to make the normalised eye images compatible between different datasets, facilitating cross-dataset evaluations.

\subsection{\methodname Architecture}
\label{subsec:CNN_model}

\begin{figure}[t]
\center
\includegraphics[width=\columnwidth]{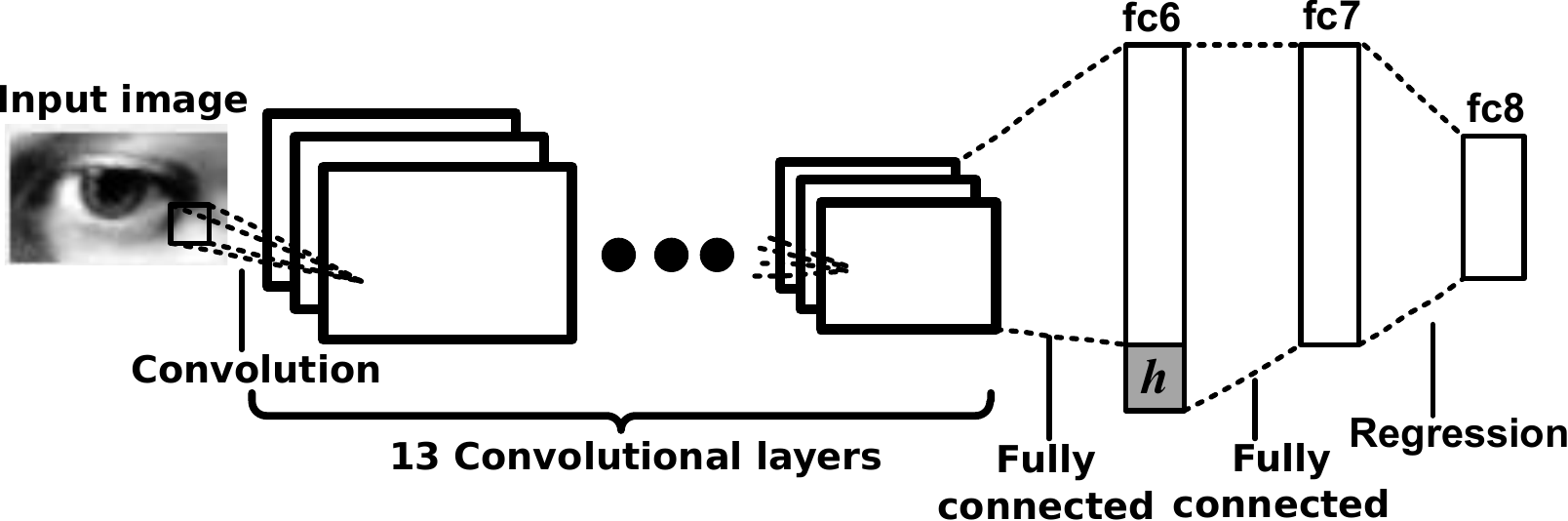}
\caption{Architecture of the proposed \methodname. The head angle $\bm{h}$ is injected into the first fully connected layer. The 13 convolutional layers are inherited from a 16-layer VGG network~\cite{Simonyan14c}.}
\label{fig:cnnmodel}
\end{figure}

The task for the CNN is to learn a mapping from the input features (2D head angle $\bm{h}$ and eye image $\bm{e}$) to gaze angles $\bm{g}$ in the normalised space.
In the unconstrained setting, the distance to the target gaze plane can vary.
The above formulation thus has the advantage that training data does not have to consider the angle of convergence between both eyes.
As pointed out in~\cite{suganolearning}, the difference between the left and right eyes is irrelevant in the person-independent evaluation scenario:
By flipping eye images horizontally and mirroring $\bm{h}$ and $\bm{g}$ around the horizontal direction, both eyes can be handled using a single regression function.

Our method is based on the 16-layer VGGNet architecture~\cite{Simonyan14c} that includes 13 convolutional layers, two fully connected layers, and one classification layer with five max pooling layers in between.
Following prior work on face~\cite{Baltrusaitis2014CCNF,chen2014joint} and gaze~\cite{sugano2015appearance,lu2014alr} analysis, we use a grey-scale single channel image as input with a resolution of $60 \times 36$ pixels.
We changed the stride of the first and second pooling layer from two to one to reflect the smaller input resolution.
The output of the network is a 2D gaze angle vector $\bm{\hat g}$ consisting of two gaze angles, yaw $\hat g_{\phi}$ and pitch $\hat g_{\theta}$.
We extended the vanilla VGGNet architecture into a multimodal model to also take advantage of head pose information~\cite{ngiam2011multimodal}.
To this end we injected head pose information $\bm{h}$ into the first fully connected layer (fc6) (see Fig.~\ref{fig:cnnmodel}).
As a loss function we used the sum of the individual ${L_2}$ losses measuring the distance between the predicted $\bm{\hat g}$ and true gaze angle vector $\bm{g}$.

\section{Experiments}

We first evaluated \methodname for cross-dataset and cross-person evaluation.
We then explored key challenges in unconstrained gaze estimation including differences in gaze ranges, illumination conditions, and personal appearance.
Finally, we studied other closely related topics, such as the influence of image resolution, the use of both eyes, and the use of head pose and pupil centre information on gaze estimation performance.
\methodname was implemented using the Caffe library~\cite{jia2014caffe}. 
We used the weights of the 16-layer VGGNet~\cite{Simonyan14c} pre-trained on ImageNet for all our evaluations, and fine-tuned the whole network in 15,000 iterations with a batch size of 256 on the training set.
We used the Adam solver~\cite{kingma2014adam} with the two momentum values set to $\beta_1 = 0.9$ and $\beta_2 = 0.95$.
An initial learning rate of 0.00001 was used and multiplied by 0.1 after every 5,000 iterations.

\subsubsection*{Baseline Methods}

We further evaluated the following baseline methods:

\begin{itemize}
\setlength\itemsep{1em}

\item \textbf{MnistNet}: 
The four-layer (two convolutional and two fully connected layers) MnistNet architecture~\cite{lecun1998gradient} has been used as the first CNN-based method for appearance-based gaze estimation~\cite{zhang2015appearance}. 
We used the implementation provided by~\cite{jia2014caffe} and trained weights from scratch. 
The learning rate was set to be 0.1 and the loss was also changed to the Euclidean distance between estimated and ground-truth gaze directions. 

\item \textbf{Random Forests (RF)}: 
Random forests were recently demonstrated to outperform existing methods for person-independent appearance-based gaze estimation~\cite{suganolearning}. 
We used the implementation provided by the authors, and the same parameters as in~\cite{suganolearning}, and we resized input eye images to $15 \times 9$ according to the implementation in~\cite{suganolearning}, which has been optimised. 

\item \textbf{$k$-Nearest Neighbours (kNN)}:
As shown in~\cite{suganolearning}, a simple kNN regression estimator can perform well in scenarios that offer a large amount of dense training images.
We used the same kNN implementation and also incorporated a training images clustering in head angle space.

\item \textbf{Adaptive Linear Regression (ALR)}:
Because it was originally designed for a person-specific and sparse set of training images~\cite{lu2014alr}, ALR does not scale well to large datasets. 
We therefore used the same approximation as in~\cite{odobez2013person}, i.e.\ we selected five training persons for each test person with lowest interpolation weights. 
We further selected random subsets of images from the neigbours of the test image in head pose space. We used the same image resolution as for RF.

\item \textbf{Support Vector Regression (SVR)}:
Schneider et al.\ used SVR with a polynomial kernel under a fixed head pose~\cite{schneider2014manifold}. 
We used a linear SVR~\cite{fan2008liblinear} for scalability given the large amount of training data. 
We also used a concatenated vector of HOG and LBP features ($6 \times 4$ blocks, $2 \times 2$ cells for HOG) as suggested in~\cite{schneider2014manifold}. 
However, we did not use manifold alignment since it does not support pose-independent training.

\item \textbf{Shape-based approach (EyeTab)}:
In addition to the appearance-based methods, we evaluated one state-of-the-art shape-based method that estimates gaze by fitting a limbus model (a fixed-diameter disc) to detected iris edges~\cite{wood14_etra}. 
We used the implementation provided by the authors.
\end{itemize}

\subsubsection*{Datasets}
\label{sec:data_preparation}

As in~\cite{zhang2015appearance}, in all experiments that follow, we used a random subset of the full dataset consisting of 1,500 left eye images and 1,500 right eye images from each participant.
Because one participant only offered 1,448 face images, we randomly oversampled data of that participant to 3,000 eye images.
From now on we refer to this subset as \emph{\datasetname}, while we call the same subset with manual facial landmark annotations \emph{\datasetnameann}.
To evaluate the generalisation capabilities of the proposed method, in addition to \datasetname, we used all screen target sequences with both VGA and HD videos of the EYEDIAP dataset for testing~\cite{FunesMora_ETRA_2014}.
We did not use the floating target sequences in the EYEDIAP dataset since they contain many extreme gaze directions that are not covered by UT Multiview.
We further used the SynthesEyes dataset~\cite{wood2015_iccv} that contains 11,382 eye samples from 10 virtual participants.

\subsubsection*{Evaluation Procedure}
For cross-dataset evaluation, each method was trained on UT Multiview or SynthesEyes, and tested on \datasetname, \datasetnameann or EYEDIAP.
We used the UT Multiview dataset as the training set for each method because it covers the largest area in head and gaze angle spaces compared to EYEDIAP and our \datasetname datasets (see Fig.\ref{fig:dataset_comparison}).
Note that SynthesEyes has the same head and gaze angle ranges as UT Multiview dataset.
For cross-person evaluation, we performed a leave-one-person-out cross-validation for all participants on \datasetnameann.

\subsection{Performance Evaluation}

We first report the performance evaluation for the cross-dataset setting, for which all the methods were trained and tested on two different datasets respectively, followed by the cross-person evaluation setting, for which all methods were evaluated with leave-one-person-out cross-validation.

\subsubsection{Cross-Dataset Evaluation}
\label{sec:cross_dataset}

\begin{figure}[t]
\center
\includegraphics[width=0.99\columnwidth]{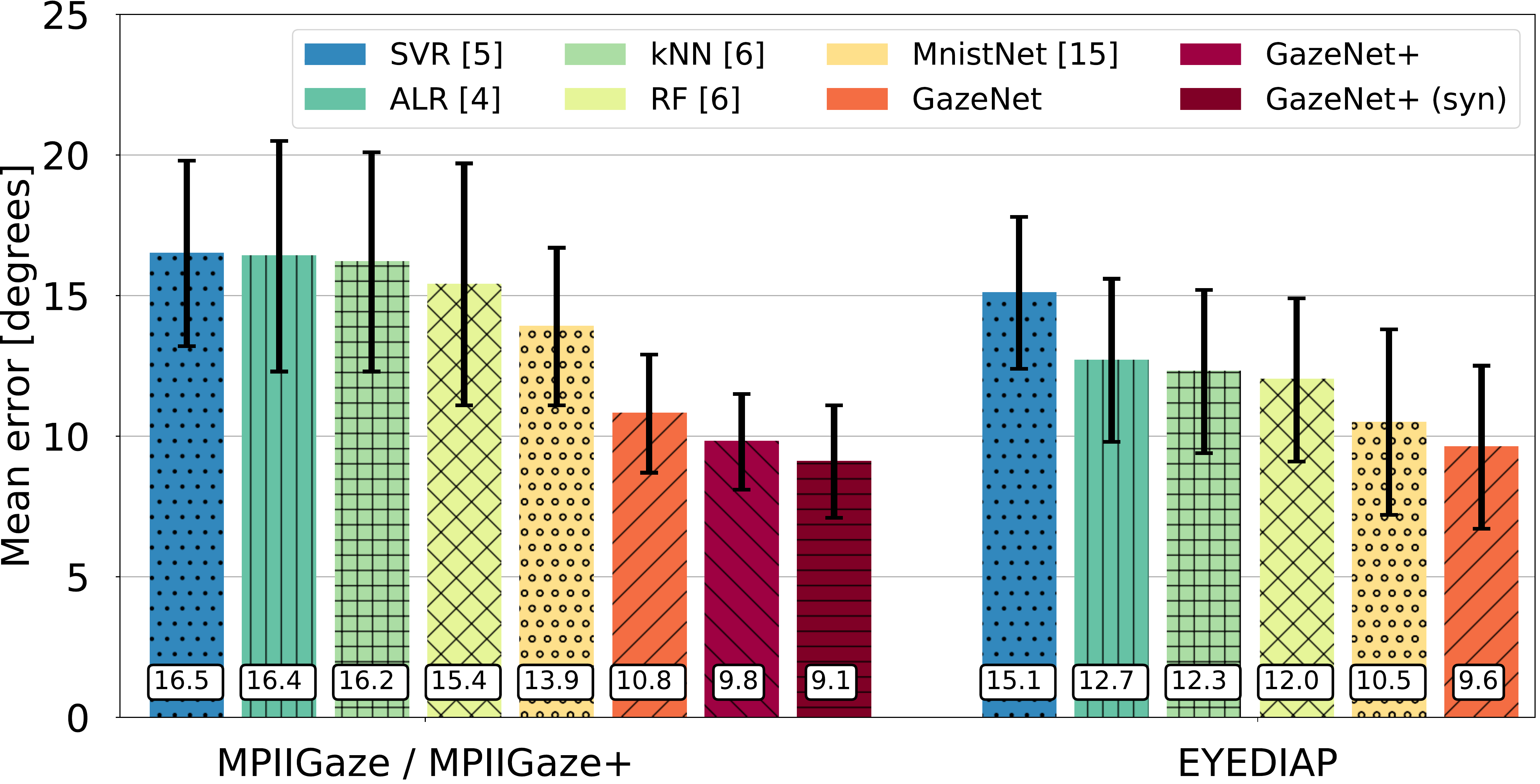}
\caption{Gaze estimation error for cross-dataset evaluation with training on 64,000 eye images in UT Multiview and testing on 45,000 eye images of \datasetname or \datasetnameann (left) and EYEDIAP (right). Bars show mean error across all participants; error bars indicate standard deviations.}
\label{fig:cross_dataset}
\end{figure}

Fig.~\ref{fig:cross_dataset} shows the mean angular errors of the different methods when trained on UT Multiview dataset and tested on both \datasetname, or \datasetnameann, and EYEDIAP datasets.
Bars correspond to mean error across all participants in each dataset, and error bars indicate standard deviations across persons.
As can be seen from the figure, our \methodname shows the lowest error on both datasets (10.8 degrees on \datasetname, 9.6 degrees on EYEDIAP). This represents a significant performance gain of 22\% (3.1 degrees) on \datasetname and 8\% on EYEDIAP (0.9 degrees), $p < 0.01$ using a paired Wilcoxon signed rank test~\cite{wilcoxon1945individual}, over the state-of-the-art method~\cite{zhang2015appearance}.
Performance on \datasetname and \datasetnameann is generally worse than on the EYEDIAP dataset, which indicates the fundamental difficulty of the in-the-wild setting covered by our dataset.
We also evaluated performance on the different sequences of EYEDIAP (not shown in the figure). Our method achieved 10.0 degrees on the HD sequences and 9.2 degrees on the VGA sequences. This difference is most likely caused by differences in camera angles and image quality.
The shape-based EyeTab method performs poorly on \datasetname (47.1 degrees mean error and 7\% misdetection rate), which shows the advantage of appearance-based approaches in this challenging cross-dataset setting.

The input image size for some baselines, like RF, kNN and ALR, has been optimized to be $15 \times 9$ pixels, which was lower than the $60 \times 36$ pixels used in our method.
To make the comparison complete, we also evaluated our \methodname with $15 \times 9$ pixels input images and achieved 11.4 degrees gaze estimation error on \datasetname, thereby still outperforming the other baseline methods.

Compared to \methodname, \methodnameann uses the manually annotated facial landmark locations \datasetnameann instead of the detected ones.
In this case the mean error is reduced from 10.8 degrees to 9.8 degrees, which indicates that the face detection and landmark alignment accuracy is still a dominant error factor in practice.
Furthermore, \methodnameann (syn) implements the strategy proposed in~\cite{wood2015_iccv}.
That is, we first trained the model with synthetic data and then fine-tuned it on the UT Multiview dataset.
This approach further reduced the gaze estimation error to 9.1 degrees.
For comparison, the naive predictor that always outputs the average gaze direction of all training eye images in UT Multiview (not shown in the figure) achieves an estimation error of 34.2 degrees on \datasetname and 42.4 degrees on EYEDIAP.

While \methodname achieved significant performance improvements for this challenging generalisation task, the results underline the difficulty of unconstrained gaze estimation. 
They also reveal a critical limitation of previous laboratory-based datasets such as UT Multiview with respect to variation in eye appearance, compared to \datasetname, which was collected in the real world.
The learning-by-synthesis approach presented in~\cite{wood2015_iccv} is promising given that it allows the synthesis of variable eye appearance and illumination conditions.
This confirms the importance of the training data and indicates that future efforts should focus on addressing the gaze estimation task both in terms of training data as well as methodology to bridge the gap to the within-dataset scenario.

\subsubsection{Cross-Person Evaluation}
\label{sec:within_dataset}

\begin{figure}[t]
\center
\includegraphics[width=0.99\columnwidth]{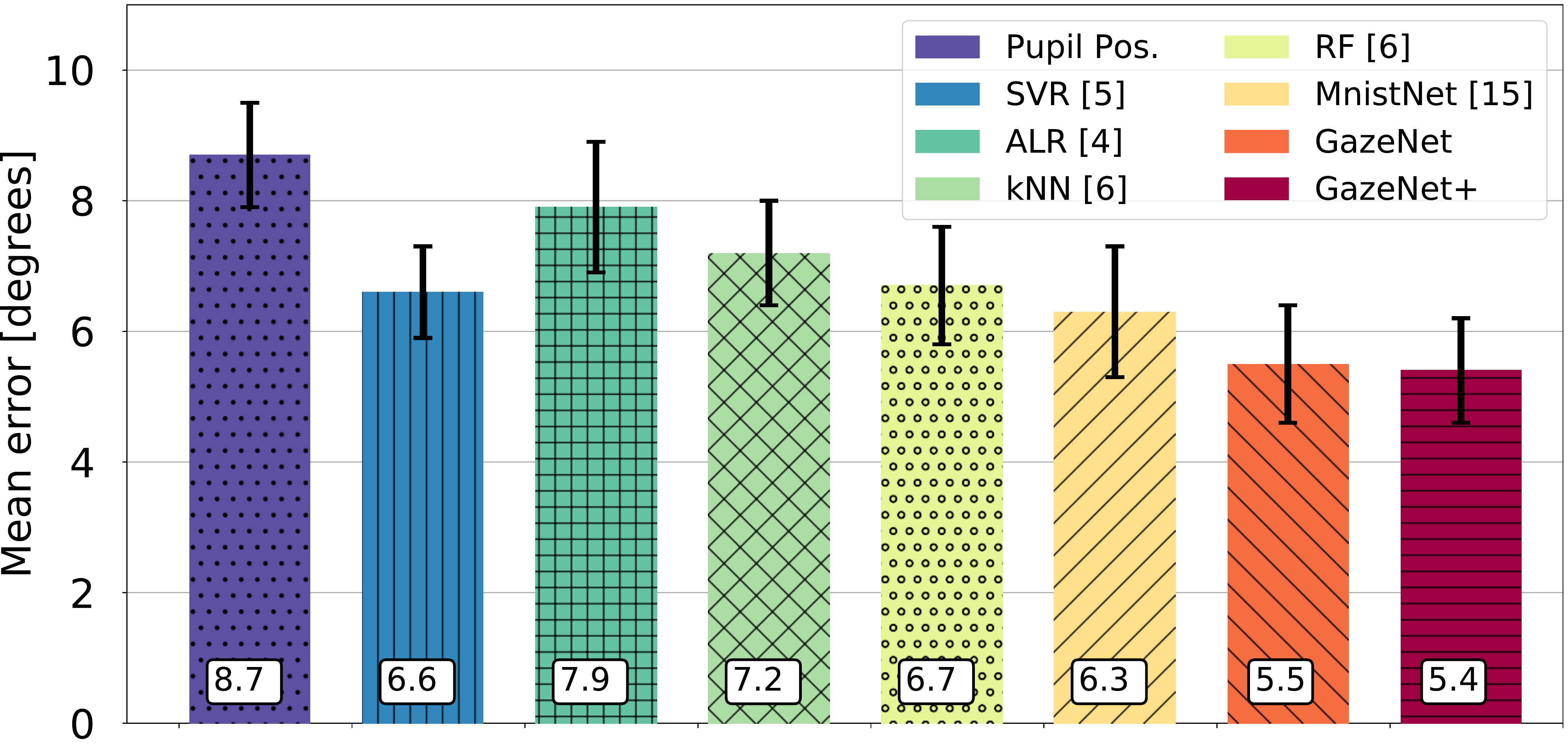}
\caption{Gaze estimation error on \datasetname and \datasetnameann for cross-person evaluation using a leave-one-person-out approach. Bars show the mean error across participants; error bars indicate standard deviations; numbers on the bottom are the mean estimation error in degrees. \methodnameann refers to the result for \datasetnameann.
}
\label{fig:within_dataset}
\end{figure}

Although results of the previous cross-dataset evaluation showed the advantage of our \methodname, they still fall short of the cross-person performance reported in~\cite{suganolearning}.
To discuss the challenges of person-independent gaze estimation within \datasetname, we performed a cross-person evaluation using a leave-one-person-out approach.
Fig.~\ref{fig:within_dataset} shows the mean angular errors of this cross-person evaluation.
Since the model-based EyeTab method has been shown to perform poorly in our setting, we opted to instead show a learning-based result using the detected pupil (iris centre) positions. 
More specifically, we used the pupil positions detected using~\cite{wood14_etra} in the normalised eye image space as a feature for kNN regression, and performed the  leave-one-person-out evaluation.

As can be seen from the figure, all methods performed better than in the cross-dataset evaluation, which indicates the importance of domain-specific training data for appearance-based gaze estimation methods.
Although the performance gain is smaller in this setting, our \methodname still significantly (13\%) outperformed the second-best MnistNet with 5.5 degrees mean error ($p < 0.01$, paired Wilcoxon signed rank test).
While the pupil position-based approach worked better than the original EyeTab method, its performance was still worse than the different appearance-based methods.
In this case there is dataset-specific prior knowledge about gaze distribution, and the mean prediction error that always outputs the average gaze direction of all training images becomes 13.9 degrees.
Because the noise in facial landmark detections is included in the training set, there was no noticeable improvement when testing our \methodname on \datasetnameann (shown as \methodnameann in Fig.~\ref{fig:within_dataset}).
It contradicts the observation with the previous cross-dataset evaluation that testing on \datasetnameann can bring one degree of improvement compared to \datasetname with detected facial landmarks (from 10.8 to 9.8 degrees).

\subsection{Key Challenges}
\label{sec:error_factor}

The previous results showed a performance gap between cross-dataset and cross-person evaluation settings.
To better understand this gap, we additionally studied several key challenges.
In all analyses that follow, we used \methodnameann in combination with \datasetnameann to minimise error in face detection and alignment.

\begin{figure}[t]
\center
\includegraphics[width=0.99\columnwidth]{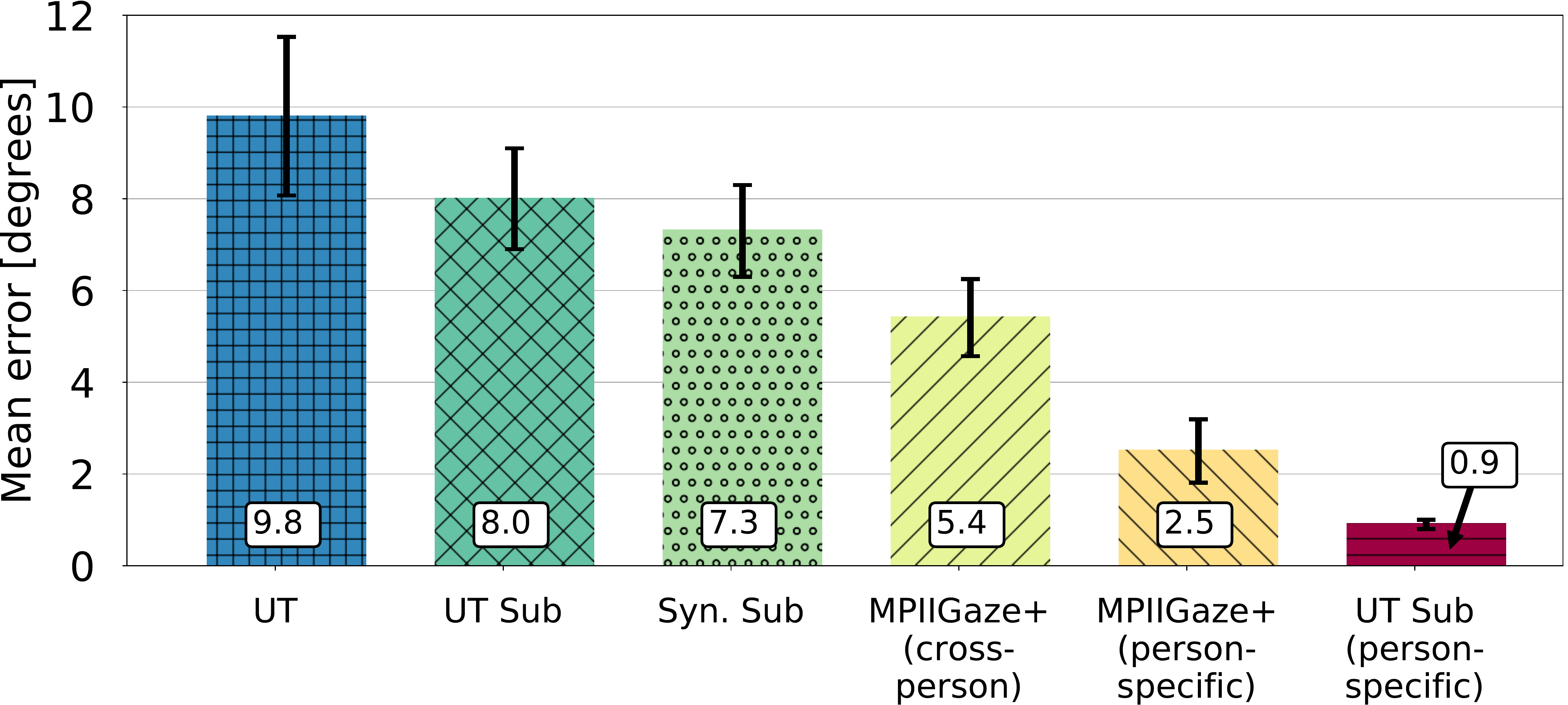}
 \caption{Gaze estimation error on \datasetnameann using \methodnameann for different training strategies and evaluation settings. Bars show the mean error across participants; error bars indicate standard deviations; numbers on the bottom are the mean estimation error in degrees. From left to right: 1) training on UT Multiview, 2) training on UT Multiview subset, 3) training on synthetic images targeted to the gaze and head pose ranges, 4) training on \datasetnameann with cross-person  evaluation, 5) training on \emph{\datasetnameann} with person-specific  evaluation, and 6) training on UT Multiview subset with person-specific evaluation.}
\label{fig:different_setting}
\end{figure}

\subsubsection{Differences in Gaze Ranges}

As discussed in~\cite{zhang2015appearance} and~\cite{wood2015_iccv}, one of the most important challenges for unconstrained gaze estimation is differences in gaze ranges between the training and testing domains.
Although handling the different gaze angles has been researched by combining geometric and appearance-based methods~\cite{mora2016gaze}, it is still challenging for appearance-based gaze estimation methods.
The first bar in Fig.~\ref{fig:different_setting} (\textit{UT}) corresponds to the cross-dataset evaluation using the UT Multiview dataset for training and \datasetnameann for testing.
In this case, as illustrated in Fig.~\ref{fig:dataset_comparison}, the training data covers wider gaze ranges than the testing data.
The second bar (\emph{UT Sub}) corresponds to the performance of the model trained on a subset of the UT Multiview dataset that consists of 3,000 eye images per participant selected so as to have the same head pose and gaze angle distributions as \datasetnameann.
If the training dataset is tailored to the target domain and the specific gaze range, we achieve about 18\% improvement in performance (from 9.8 to 8.0 degrees).

\begin{figure}[t]
\center
\includegraphics[width=0.99\columnwidth]{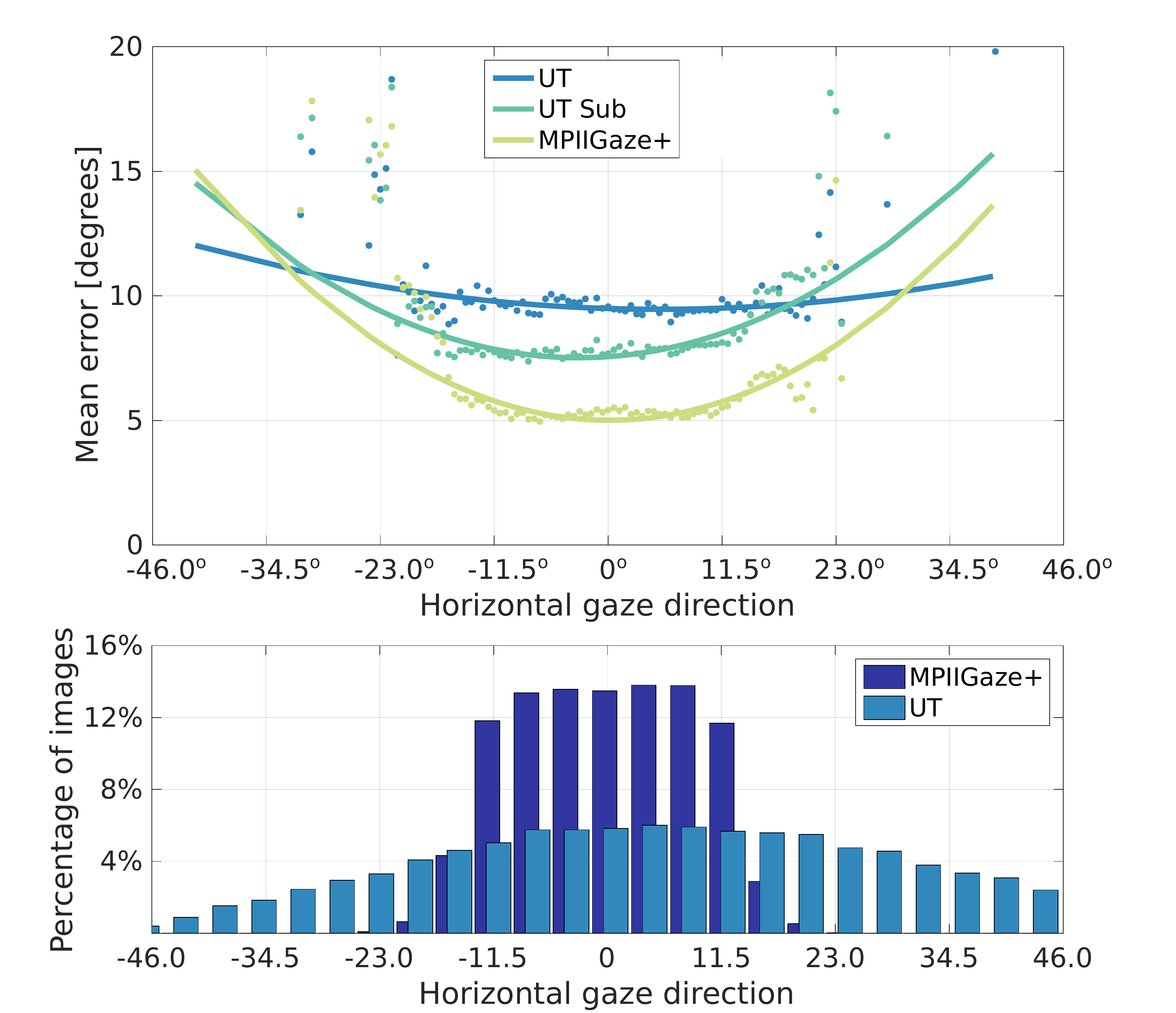}
\caption{Top: Gaze estimation error on \datasetnameann for the model trained with UT Multiview, UT Multiview subset, and \datasetnameann for different horizontal gaze directions. Bottom: 
Percentage of images for the horizontal gaze directions of \datasetnameann and UT.}
\label{fig:performance_gaze_range}
\end{figure}

The top of Fig.~\ref{fig:performance_gaze_range} shows the gaze estimation errors in horizontal gaze direction with training on UT Multiview, UT Multiview subset, and \datasetnameann, and testing on \datasetnameann.
The dots correspond to the average error for that particular gaze direction, while the line is the result of a quadratic polynomial curve fitting.
The lines correspond to the \emph{UT}, \emph{UT Sub} and \emph{\datasetnameann (cross-person)} bars in Fig.~\ref{fig:different_setting}.
As can be seen from the figure, for the model trained on UT Multiview subset, gaze estimation error increased for images that were close to the edge of the gaze range.
In contrast, the model trained on the whole UT Multiview showed more robust performance across the full gaze direction range.
The most likely reason for this difference is given by Fig.~\ref{fig:performance_gaze_range}, which showns the percentage of images for the horizontal gaze directions for the training samples of \datasetnameann and UT Multiview.
As can be seen from the figure, while \emph{UT Sub} and \datasetnameann have the same gaze direction distribution, UT Multiview and \datasetnameann differ substantially.
This finding demonstrates the fundamental shortcoming of previous works that only focused on cross-person evaluations and thereby implicitly or explicitly assumed a single, and thus restricted, gaze range.
As such, this finding highlights the importance not only of cross-dataset evaluations, but also of developing methods that are robust to (potentially very) different gaze ranges found in different settings.

\subsubsection{Differences in Illumination Conditions}

Illumination conditions are another important factor in unconstrained gaze estimation and have been the main motivation for using fully synthetic training data that can cover a wider range of different illuminations~\cite{wood2015_iccv}.
The third bar in Fig.~\ref{fig:different_setting} (\emph{Syn Sub}) corresponds to the same fine-tuned model as \emph{\methodnameann (syn)} in Fig.~\ref{fig:cross_dataset}, but with the gaze range restricted to the same head pose and gaze angle distributions as \datasetnameann.
The fourth bar in Fig.~\ref{fig:different_setting} (\emph{\datasetnameann (cross-person)}) shows the results of within-dataset cross-person evaluation on~\datasetnameann.
For the second to the fourth bar in Fig.~\ref{fig:different_setting}, the training data has nearly the same head angle and gaze direction range.
The only difference is in the variation in illumination conditions in the training data.
While the use of synthetic training data results in improved performance (from 8.0 degrees to 7.3 degrees), 
there is still a large gap between cross-dataset and cross-person settings.

\begin{figure}[t]
\center
\includegraphics[width=0.99\columnwidth]{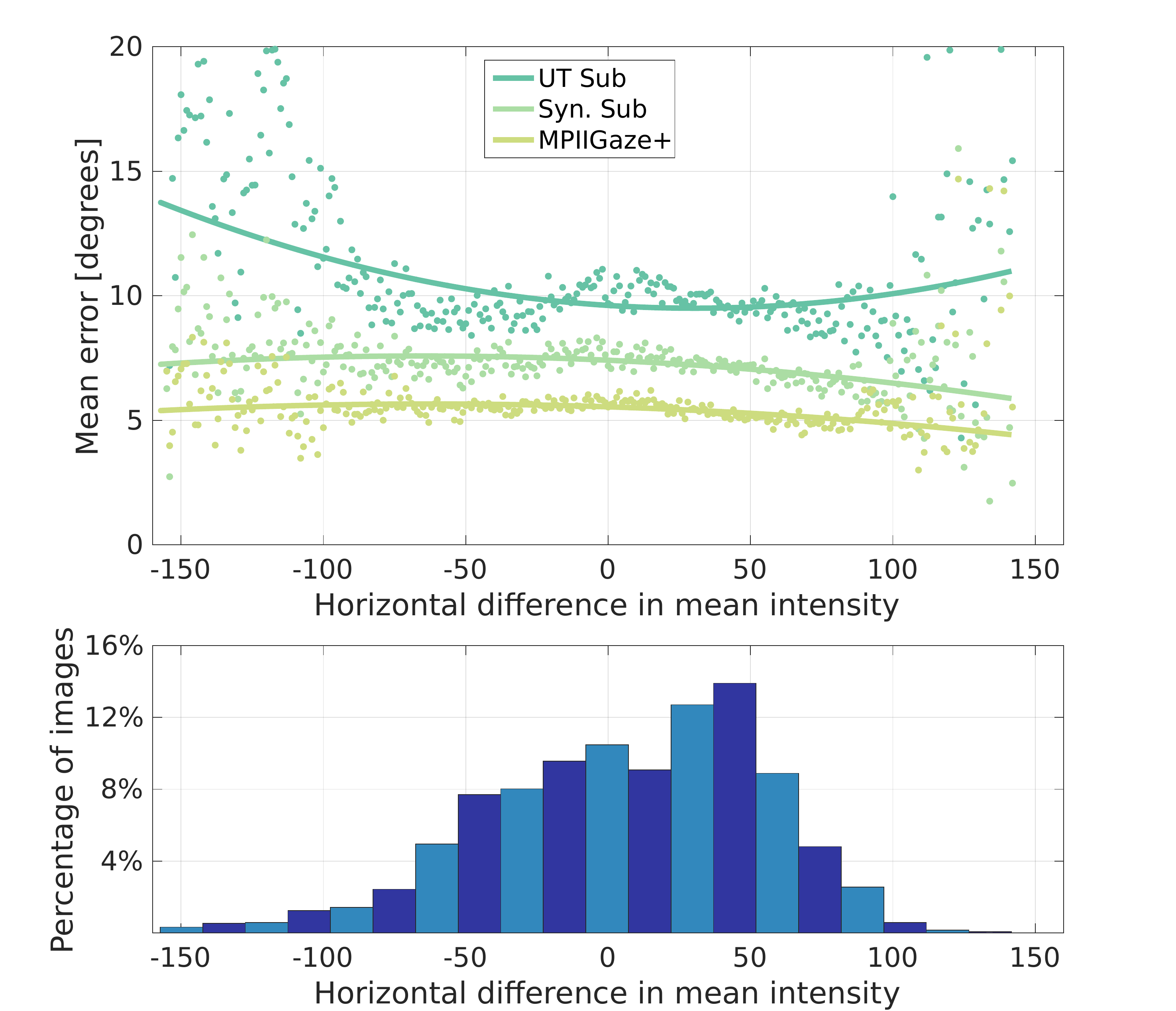}
\caption{Top: Gaze estimation error on \datasetnameann across mean greyscale intensity differences between the left and right half of the face region for models trained on UT subset, SynthesEyes subset, and \datasetnameann. Bottom: Corresponding percentage of images for all mean greyscale intensity differences.
}
\label{fig:performance_directional_light}
\end{figure}

This tendency is further illustrated in Fig.~\ref{fig:performance_directional_light}, in which we evaluated gaze estimation error with respect to lighting directions with our \methodname.
Similar to Fig.~\ref{fig:characteristics}, we plotted the mean gaze estimation error according to the mean intensity difference between the left and right face region.
The different colours represent the models trained with UT Multiview subset, synthetic subset and \datasetnameann.
They also correspond to \emph{UT Sub}, \emph{Syn. Sub} and \emph{\datasetnameann (cross-person)} in Fig.~\ref{fig:different_setting}.
The dots are averaged error for horizontal difference in the mean intensity in the face region, and lines are with quadratic polynomial curve fitting.
Similar to Fig.~\ref{fig:performance_gaze_range}, the bottom of Fig.~\ref{fig:performance_directional_light} shows the percentage of images for mean greyscale intensity difference between the left and right half of the face region.
We cannot show the distribution for \emph{UT Sub} and \emph{Syn. Sub} since their face images are not available.
Compared to the model trained solely on the UT subset, the model with synthetic data shows better performance across different lighting conditions. 
While there still remains an overall performance gap from the domain-specific performance, the effect of synthetic data is more visible in the area with extreme lighting directions.

\subsubsection{Differences in Personal Appearance}

To further study the unconstrained gaze estimation task, we then evaluated person-specific gaze estimation performance, i.e.\ where training and testing data come from the same person.
The results of this evaluation on \datasetnameann are shown as the second last bar (\emph{\datasetnameann (person-specific)}) in Fig.~\ref{fig:different_setting}. 
Since there are 3,000 eye images for each participant in \datasetnameann, we picked the first 2,500 eye images for training and the rest for testing. 
Similarly, the last bar ({\em UT Sub (p.s.)}) in Fig.~\ref{fig:different_setting} shows the person-specific evaluation within the UT subset, also with 2,500 eye images for training and 500 eye images for testing.
The performance gap between \emph{\datasetnameann (cross-person)} and \emph{\datasetnameann (person-specific)} illustrates the fundamental difficulty of person-independent gaze estimation.
The difference between \emph{\datasetnameann} (person-specific) and \emph{UT Sub (p.s.)} also shows, however, that in-the-wild settings are challenging even for the person-specific case.

\begin{figure}[t]
\center
\begin{subfigure}[r]{\linewidth}
	\centering
	\includegraphics[width=\textwidth]{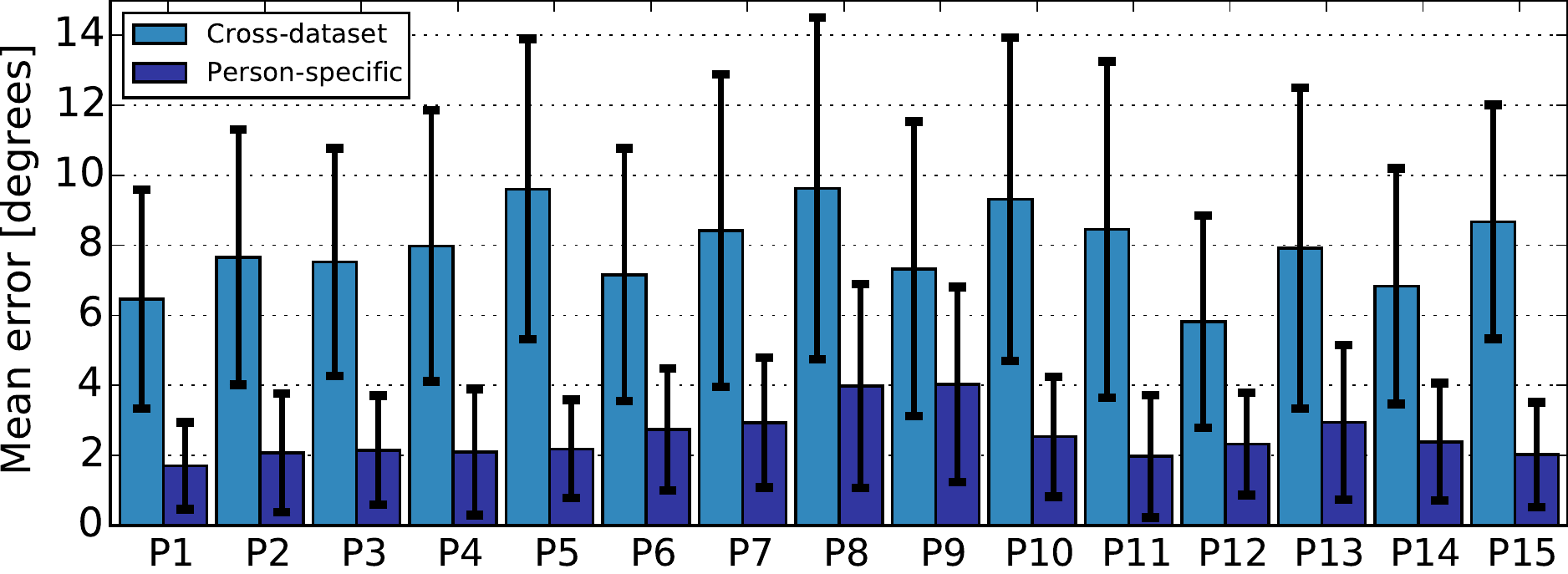}
\end{subfigure}\\
\vspace{0.1cm}
\hspace{0.5cm}
\begin{subfigure}[b]{0.062\linewidth}
	\centering
	\includegraphics[width=\textwidth]{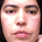}
\end{subfigure}%
\begin{subfigure}[b]{0.062\linewidth}
	\centering
	\includegraphics[width=\textwidth]{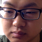}
\end{subfigure}%
\begin{subfigure}[b]{0.062\linewidth}
	\centering
	\includegraphics[width=\textwidth]{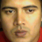}
\end{subfigure}%
\begin{subfigure}[b]{0.062\linewidth}
	\centering
	\includegraphics[width=\textwidth]{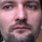}
\end{subfigure}%
\begin{subfigure}[b]{0.062\linewidth}
	\centering
	\includegraphics[width=\textwidth]{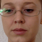}
\end{subfigure}%
\begin{subfigure}[b]{0.062\linewidth}
	\centering
	\includegraphics[width=\textwidth]{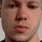}
\end{subfigure}%
\begin{subfigure}[b]{0.062\linewidth}
	\centering
	\includegraphics[width=\textwidth]{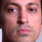}
\end{subfigure}%
\begin{subfigure}[b]{0.062\linewidth}
	\centering
	\includegraphics[width=\textwidth]{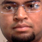}
\end{subfigure}%
\begin{subfigure}[b]{0.062\linewidth}
	\centering
	\includegraphics[width=\textwidth]{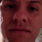}
\end{subfigure}%
\begin{subfigure}[b]{0.062\linewidth}
	\centering
	\includegraphics[width=\textwidth]{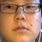}
\end{subfigure}%
\begin{subfigure}[b]{0.062\linewidth}
	\centering
	\includegraphics[width=\textwidth]{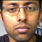}
\end{subfigure}%
\begin{subfigure}[b]{0.062\linewidth}
	\centering
	\includegraphics[width=\textwidth]{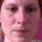}
\end{subfigure}%
\begin{subfigure}[b]{0.062\linewidth}
	\centering
	\includegraphics[width=\textwidth]{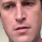}
\end{subfigure}%
\begin{subfigure}[b]{0.062\linewidth}
	\centering
	\includegraphics[width=\textwidth]{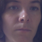}
\end{subfigure}%
\begin{subfigure}[b]{0.062\linewidth}
	\centering
	\includegraphics[width=\textwidth]{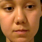}
\end{subfigure}%

\caption{Gaze estimation error for each participant for two evaluation schemes: \emph{cross-dataset}, where the model was trained on UT Multiview and tested on \datasetnameann, and \emph{person-specific}, where the model was trained and tested on the same person from \datasetnameann. Sample images are shown at the bottom.}
\label{fig:performance_participant}
\end{figure}

Fig.~\ref{fig:performance_participant} shows the estimation error of each participant in both cross-dataset (trained on the UT Multiview) and person-specific (leave-one-person-out training on \datasetnameann) settings with our \methodname.
Bars correspond to mean error for each participant and the error bars indicate standard deviations. 
Example faces from each participant are shown at the bottom.
As the figure shows, for the cross-dataset evaluation the worst performance was achieved for participants wearing glasses (P5, P8, and P10). 
This is because the UT Multiview dataset does not include training images covering this case, although glasses can cause noise in the eye appearance as shown in Fig.~\ref{fig:dataset_comparison:mpii-images-glasses}.
For the person-specific evaluation, glasses are not the biggest error source, given that corresponding images are available in the training set. 
It can also be seen that the performance differences between participants are smaller in the person-specific evaluation.
This indicates a clear need for developing new methods that can robustly handle differences in personal appearance for unconstrained gaze estimation.

\subsection{Further Analyses}
\label{sec:model_training}

Following the previous evaluations of unconstrained gaze estimation performance and key challenges, we now provide further analyses on closely related topics, specifically the influence of image resolution, the use of both eyes, and the use of head pose and pupil centre information on gaze estimation performance.

\subsubsection{Image Resolution}
\label{sec:evaluation_resolution}

\begin{figure}[t]
\begin{subfigure}[r]{0.40\linewidth}
	\centering
	\includegraphics[width=\textwidth]{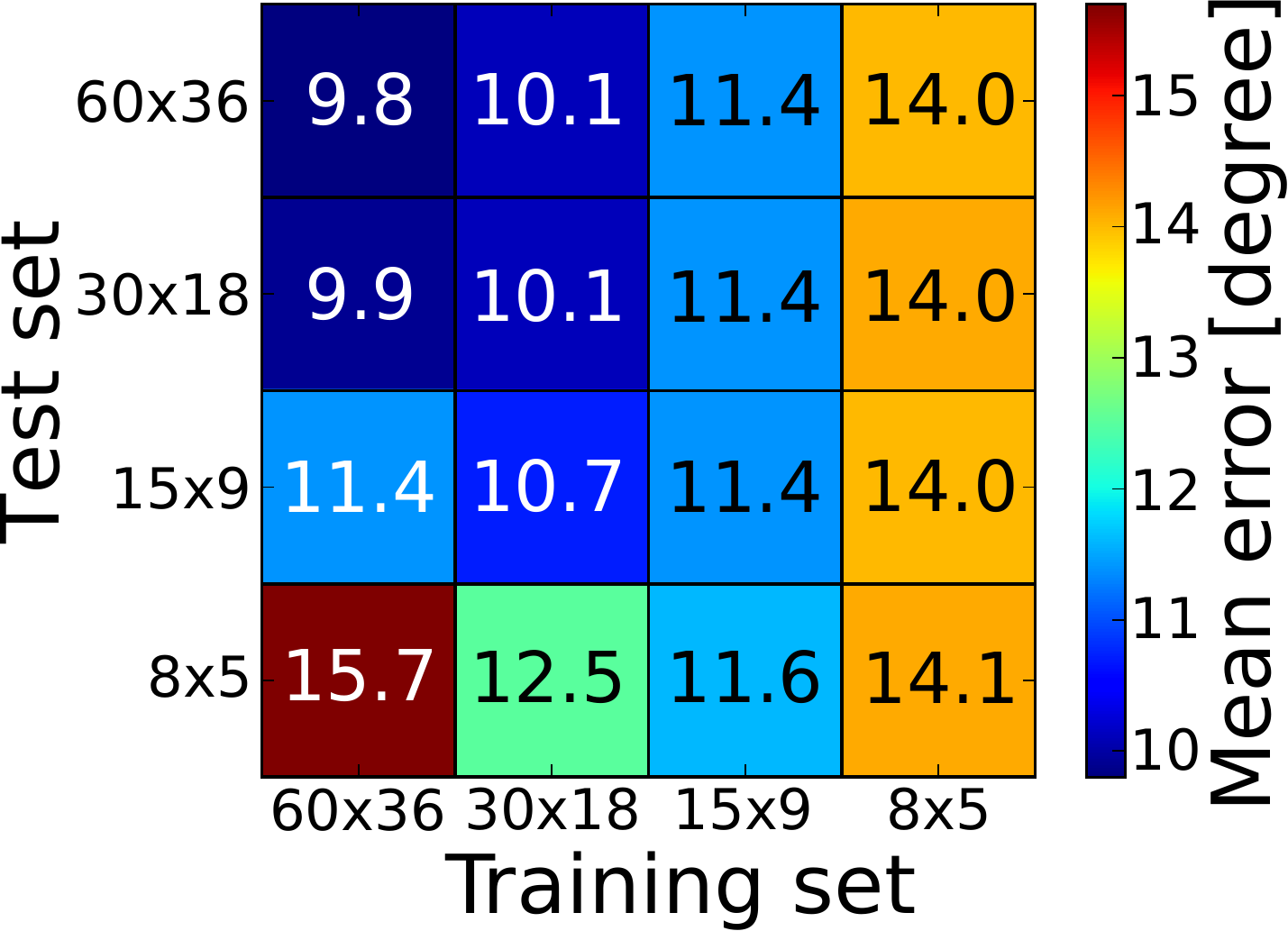}
	\caption{}
    \label{fig:performace_resolution_relative}
\end{subfigure}
~
\begin{subfigure}[r]{0.60\linewidth}
	\centering
	\includegraphics[width=\textwidth]{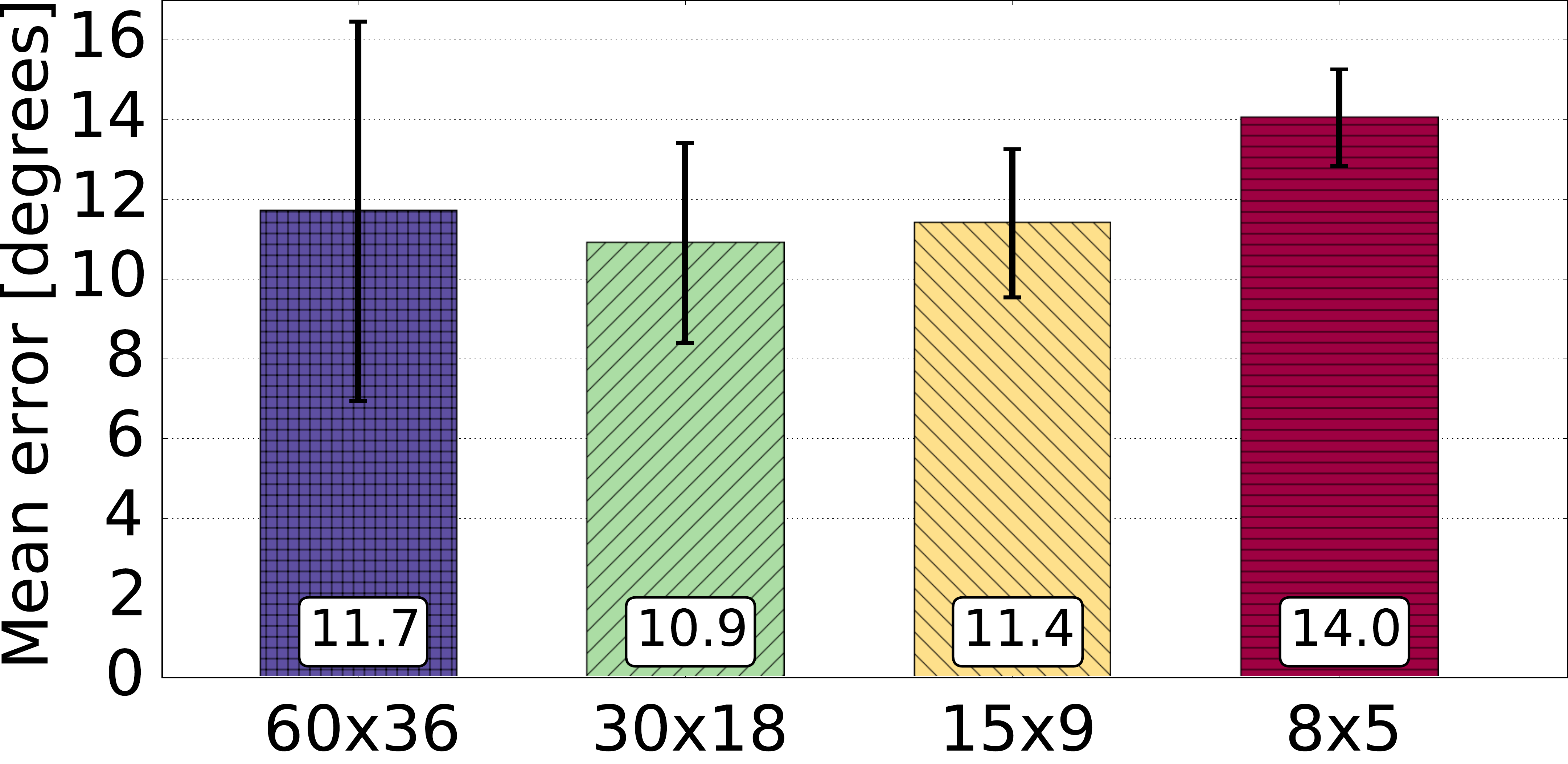}
	\caption{}
    \label{fig:performace_resolution_his}
\end{subfigure}
\caption{Gaze estimation error of the models trained on UT Multiview and tested on \datasetnameann for different image resolutions. Test images were resized to the resolution of the training images. (a) Combinations of different training and test set resolutions with cell numbers indicating the average error in degrees. (b) The mean estimation error for the models trained with certain image resolutions across all images. Bars show the mean error across participants in degrees; error bars indicate standard deviations.}
\label{fig:performace_resolution}
\end{figure}

We first explored the influence of image resolution on gaze estimation performance, since it is conceivable that this represents a challenge for unconstrained gaze estimation.
To this end, we evaluated the performance for the cross-dataset evaluation setting (trained on UT Multiview and tested on \datasetnameann) for different training and testing resolutions with our \methodname.
Starting from the default input resolution $60\times36$ in our model, we reduced the size to $30\times18$, $15\times9$ and $8\times5$.
We always resized the test images according to the training resolution with bicubic interpolation.
During training, we modified the stride of the first convolutional and max pooling layers of our \methodname accordingly so that the input became the same starting from the second convolutional layer, regardless of the original image input resolution.
Fig.~\ref{fig:performace_resolution_relative} summarises the results of this evaluation with resolutions of training images along the x-axis, and resolutions of testing images on the y-axis.
In general, if the test images have higher resolution than the training images, higher resolution results in better performance.
Performance becomes significantly worse if the test images are smaller than the training images.

Fig.~\ref{fig:performace_resolution_his} shows the mean error of these models trained on one image resolution and tested across all testing resolutions, with the error bar denoting the standard deviation across all images.
For the reason discussed above, the overall performance of the highest-resolution model is worse than that of the second $30 \times 18$ model.
This shows that higher resolution does not always mean better performance for unconstrained gaze estimation.

\subsubsection{Use of Both Eyes}
\label{sec:evaluation_oracle}

Previous methods typically used a single eye image as input.
However, it is reasonable to assume that for some cases, such as strong directional lighting, performance can be improved by using information from both eyes.
To study this in more detail, we selected all images from \datasetnameann with two annotated eyes.
We then evaluated different means of merging information from both eyes.
The gaze estimation error when averaging across both eye images using the model trained on the UT Multiview dataset is 9.8 degrees with a standard deviation of 2.1 degrees.
The best-case performance, i.e.\ always selecting the eye showing lower gaze estimation error, is 8.4 degrees with a standard deviation of 1.9 degrees.
The gap between these two bars illustrates the limitations of the single eye-based estimation approach.

One approach to integrate estimation results from both eyes is to geometrically merge 3D gaze vectors after the appearance-based estimation pipeline.
Given two 3D gaze vectors from both eyes, we thus further computed the mean gaze vector originating from the centre of both eyes.
Ground-truth gaze vectors were also defined from the same origin, and the mean error across all faces using this approach was 7.2 degrees (standard deviation 1.4 degrees).
It can be seen that even such a simple late fusion approach improves the estimation performance, indicating the potential of more sophisticated methods for fusing information from both eyes. 

\subsubsection{Use of Head Pose Information}
\label{sec:evaluation_headpose}

\begin{figure}[t]
\center
\includegraphics[width=0.99\columnwidth]{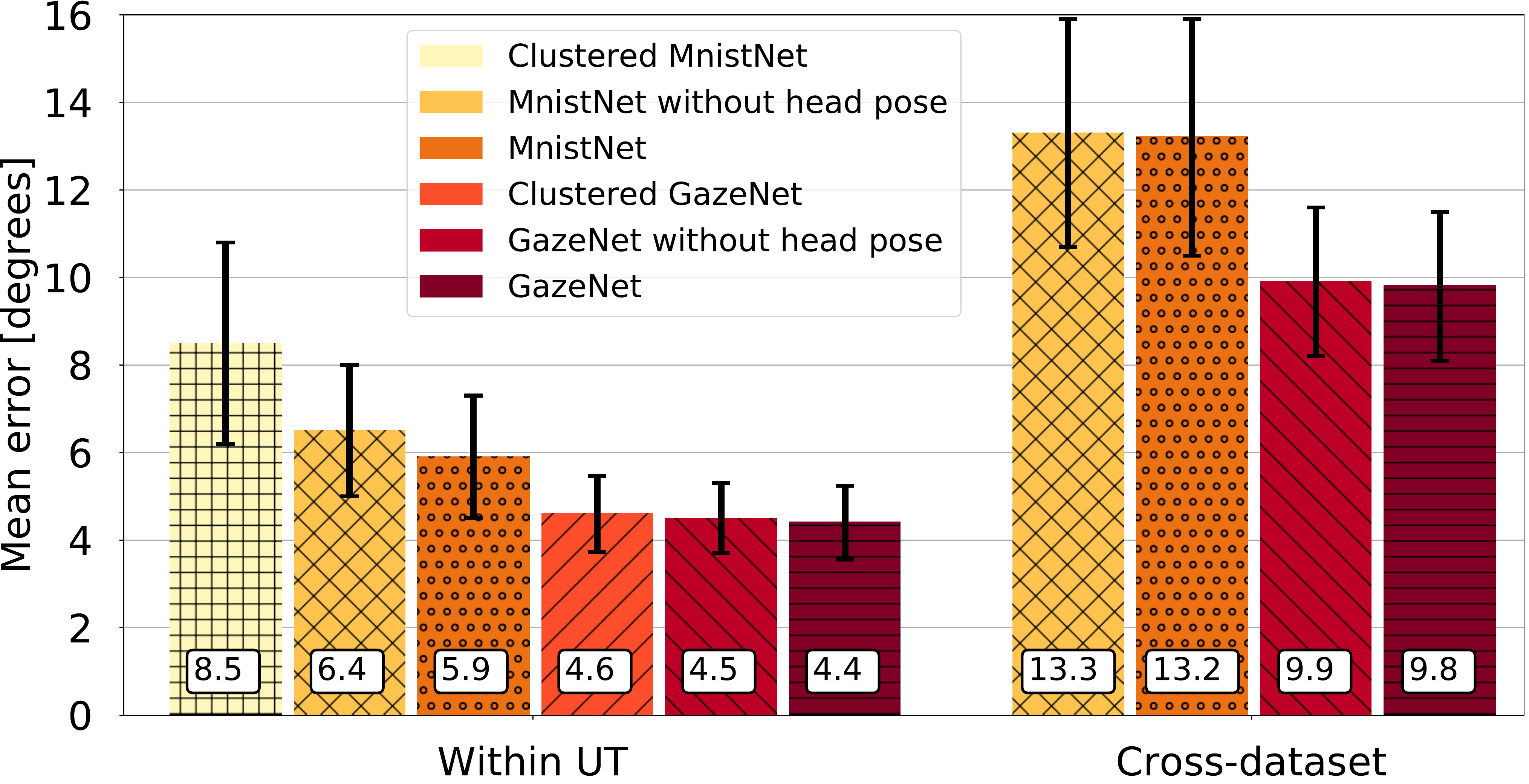}
\caption{Gaze estimation error when using the pose-clustered structure (\emph{Clustered MnistNet} and \emph{Clustered \methodname}), without head angle vectors $\bm{h}$ (\emph{MnistNet without head pose} and \emph{\methodname without head pose}) for within-UT and cross-dataset (trained on UT, tested on \datasetnameann) settings. Bars show the mean error across participants; error bars indicate standard deviations; numbers on the bottom are the mean estimation error in degrees.}
\label{fig:performance_headpose}
\end{figure}

To handle arbitrary head poses in the gaze estimation task, 3D head pose information has been used for the data normalisation as described in Sec.~\ref{subsec:normalisation}.
After normalisation, 2D head angle vectors $\bm{h}$ were injected into the network as an additional geometry feature.
The left side of Fig.~\ref{fig:performance_headpose} shows a comparison between different architectures of the multi-modal CNN on the UT Multiview dataset.
We followed the same three-fold cross-validation setting as in~\cite{suganolearning}. 
The best performance reported in~\cite{suganolearning} is 6.5 degrees mean estimation error achieved by the head pose-clustered Random Forest.
However, when the same clustering architecture is applied to the MnistNet ({\em Clustered MnistNet}), the performance became worse than for the model without clustering.
In addition, our \methodname ({\em Clustered \methodname}) did not show any noticeable difference with the clustering structure.
This indicates the higher learning flexibility of the CNN, which contributed to the large performance gain in the estimation task.
The role of the additional head pose feature is also different in the two CNN architectures.
While the MnistNet architecture achieved better performance with the help of the head pose feature, the effect of the head pose feature became marginal in the case of the \methodname.
Even though deeper networks like \methodname can in general achieve better performance, achieving better performance with shallower networks is still important in some practical use cases where tehre is limited computational power, such as on mobile devices.

The right side of Fig.~\ref{fig:performance_headpose} shows a comparison of models with and without the head pose feature in the cross-dataset setting (trained on UT and tested on \datasetnameann).
The effect of the additional head pose feature is marginal in this case, but this is likely because the head pose variation in the \datasetname dataset is already limited to near-frontal cases.
We performed an additional experiment to compare the gaze estimation performance when using the head pose estimated from the personal and the generic 3D face model.
We achieved 9.8 degrees and 9.7 degrees for the cross-dataset evaluation, respectively, suggesting that the generic face model is sufficiently accurate for the gaze estimation task.

\subsubsection{Use of Pupil Centres}
\label{sec:evaluation_pupil}
In \methodname, we do not use pupil centre information as input.
Although intuitively, eye shape features, such as pupil centres, can be a strong cue for gaze estimation, the model- or shape-based baseline performed relatively poorly for both the cross-dataset and cross-person evaluations.
We therefore finally evaluated the performance of \methodname when using the pupil centre as an additional feature for cross-person evaluation on \datasetnameann.
We detected the pupil centre location inside the normalised eye images using~\cite{wood14_etra} and concatenated the pupil location to the geometry feature vector (head angle $\bm{h}$).
While there was an improvement between the models without and with the pupil centre feature, the improvement was relatively small (from 5.4 to 5.2 degrees).
Performance improved more when using the manually annotated pupil centres, but still not significantly (5.0 degrees).

\section{Discussion}

This work made an important step towards unconstrained gaze estimation, i.e.\ gaze estimation from a single monocular RGB camera without assumptions regarding users' facial appearance, geometric properties of the environment or the camera and user therein.
Unconstrained gaze estimation represents the practically most relevant but also most challenging gaze estimation task.
Unconstrained gaze estimation is, for example, required for second-person gaze estimation from egocentric cameras or by a mobile robot.
Through cross-dataset evaluation on our new \datasetname dataset, we demonstrated the fundamental difficulty of this task compared to the commonly used person-independent, yet still domain-specific, evaluation scheme.
Specifically, gaze estimation performance dropped by up to 69\% (from a gaze estimation error of 5.4 to 9.1 degrees) for the cross-dataset evaluation, as can be seen by comparing Figs.~\ref{fig:cross_dataset} and \ref{fig:within_dataset}.
The proposed \methodname significantly outperformed the state of the art for both evaluation settings and in particular when pre-trained on synthetic data (see Fig.~\ref{fig:cross_dataset}).
The 3.1 degrees improvement that we achieved in the cross-dataset evaluation corresponds to around 2.9 cm on the laptop screen after backprojection.
Performance on~\datasetname was generally worse than on EYEDIAP, which highlights the difficulty but also the importance of developing and evaluating gaze estimators on images collected in real-world environments.

We further explored key challenges of this task, including differences in gaze ranges, illumination conditions, and personal appearance.
Previous works either implicitly or explicitly side stepped these challenges by restricting the gaze or head pose range~\cite{McMurrough:2012:ETD:2168556.2168622,Ponz:2012:DEE:2370216.2370364}, studying a fixed illumination condition~\cite{he2015omeg,FunesMora_ETRA_2014,suganolearning}, or by only recording for short amounts of time and thereby limiting variations in personal appearance~\cite{smith2013gaze,weidenbacher07}.
Several recent works also did not study 3D gaze estimation but, instead, simplified the task to regression from eye images to 2D on-screen coordinates~\cite{Huang2017,krafka2016eye}.
While the 3D gaze estimation task generalises across hardware and geometric settings and thus facilities full comparison with other methods, the 2D task depends on the camera-screen relationship.
Our evaluations demonstrated the fundamental shortcomings of such simplifications.
They also showed that the development of 3D gaze estimation methods that properly handle all of these challenges, while important, remains largely unexplored.
The ultimate goal of unconstrained gaze estimation is to obtain a generic estimator that can be distributed as a pre-trained library.
While it is challenging to learn estimators that are robust and accurate across multiple domains, an intermediate solution might be to develop methods that adapt using domain-specific data automatically collected during deployment~\cite{sugano16_uist,zhang17_uist}.

The head angle vector plays different roles for the cross- and within-dataset evaluations.
It is important to note that a 3D formulation is always required for unconstrained gaze estimation without restricting the focal length of the camera or the pose of the gaze target plane.
3D geometry, including the head pose, therefore has to be handled properly for unconstrained gaze estimation -- a challenge still open at the moment.
In this work we additionally explored the use of the head angle vector as a separate input to the CNN architecture as described in~\cite{zhang2015appearance}.
As shown in Fig.~\ref{fig:performance_headpose}, while head pose information does result in a performance improvement for the shallower MnistNet architecture used in~\cite{zhang2015appearance}, it does not significantly improve the performance of~\methodname.

The state-of-the-art shape-based method~\cite{wood14_etra} performed poorly in the cross-dataset evaluation, achieving only 47.1 degrees mean error.
Similarly, adding the detected pupil centres as additional input to the CNN resulted in only a small performance improvement (see Section~\ref{sec:evaluation_pupil}).
While using eye shape and pupil centre features is typically considered to be a promising approach, both findings suggest that its usefulness may be limited for unconstrained gaze estimation, particularly on images collected in real-world settings -- leaving aside the challenge of detecting these features robustly and accurately on such images in the first place.

\section{Conclusion}
\label{sec:conclusion}

In this work we made a case for unconstrained gaze estimation -- a task that, despite its scientific and practical importance, has been simplified in several ways in prior work.
To address some of these simplifications, we presented the new \datasetname dataset that we collected over several months in everyday life and that therefore covers significant variation in eye appearance and illumination.
The dataset also offers manually annotated facial landmarks for a large subset of images and is therefore well-suited for cross-dataset evaluations.
Through extensive evaluation of several state-of-the-art appearance- and model-based gaze estimation methods, we demonstrated both the critical need for and challenges of developing new methods for unconstrained gaze estimation.
Finally, we proposed an appearance-based method based on a deep convolutional neural network that improves performance by 22\% for the most challenging cross-dataset evaluation on \datasetname.
Taken together, our evaluations provide a detailed account of the state of the art in appearance-based gaze estimation and guide future research on this important computer vision task.

\ifCLASSOPTIONcompsoc
  \section*{Acknowledgments}
\else
  \section*{Acknowledgment}
\fi

We would like to thank Laura Sesma for her help with the dataset handling and normalisation.
This work was funded, in part, by the Cluster of Excellence on Multimodal Computing and Interaction (MMCI) at Saarland University, Germany, an Alexander von Humboldt Postdoctoral Fellowship, Germany, and a JST CREST Research Grant (JPMJCR14E1), Japan.

\ifCLASSOPTIONcaptionsoff
  \newpage
\fi

{
\bibliographystyle{IEEEtran}
\bibliography{references}
}

\vspace{-1cm}

\begin{IEEEbiography}[{\includegraphics[width=1in,height=1.25in,clip,keepaspectratio]{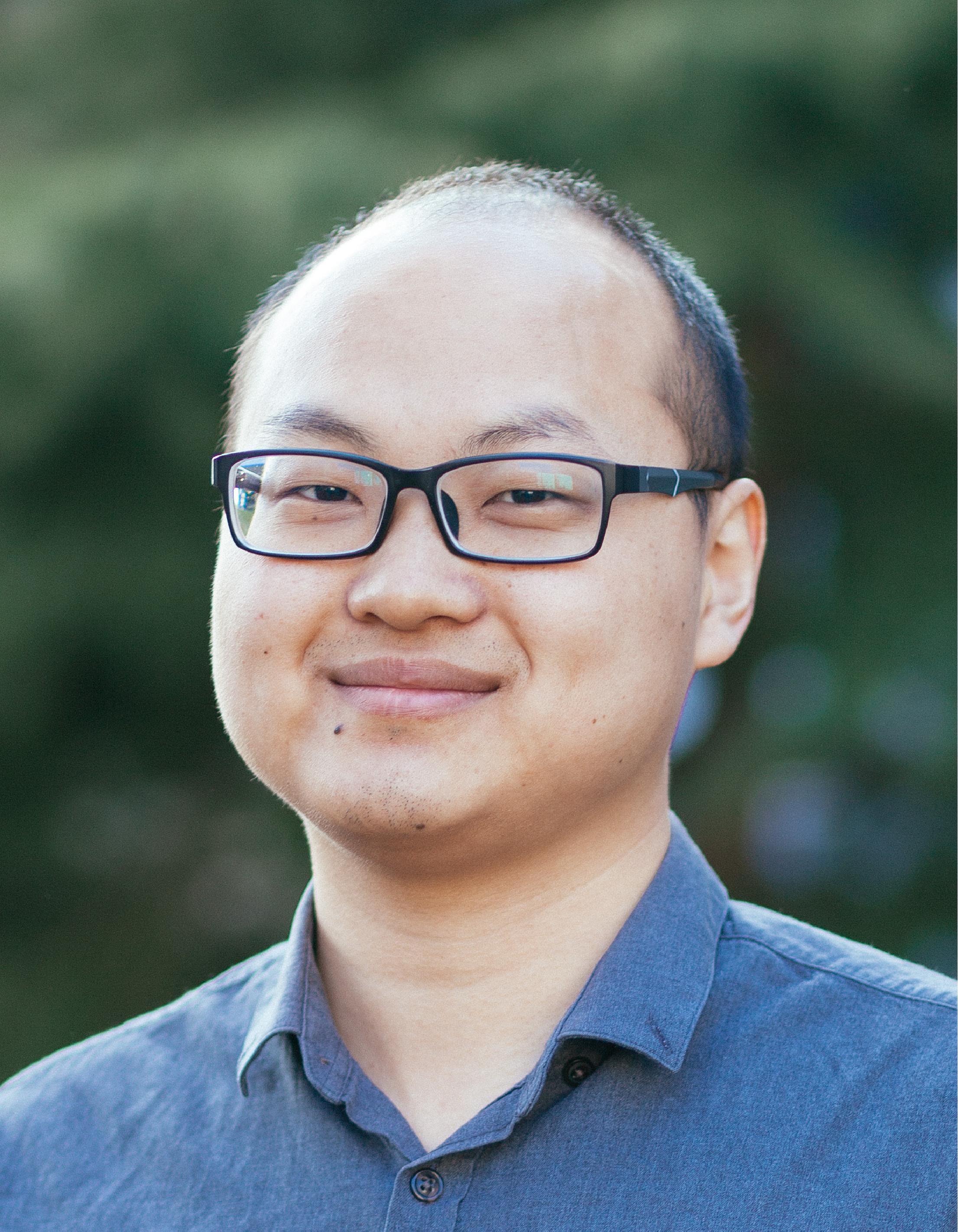}}]{Xucong Zhang}
is a PhD student in the Perceptual User Interfaces Group at the Max Planck Institute for Informatics, Germany. Xucong Zhang received his BSc. from the China Agriculture University in 2007, and a MSc. from Beihang University in 2010. His research interests include computer vision, human-computer interaction, and learning-based gaze estimation.
\end{IEEEbiography}

\vspace{-1cm}

\begin{IEEEbiography}[{\includegraphics[width=1in,height=1.25in,clip,keepaspectratio]{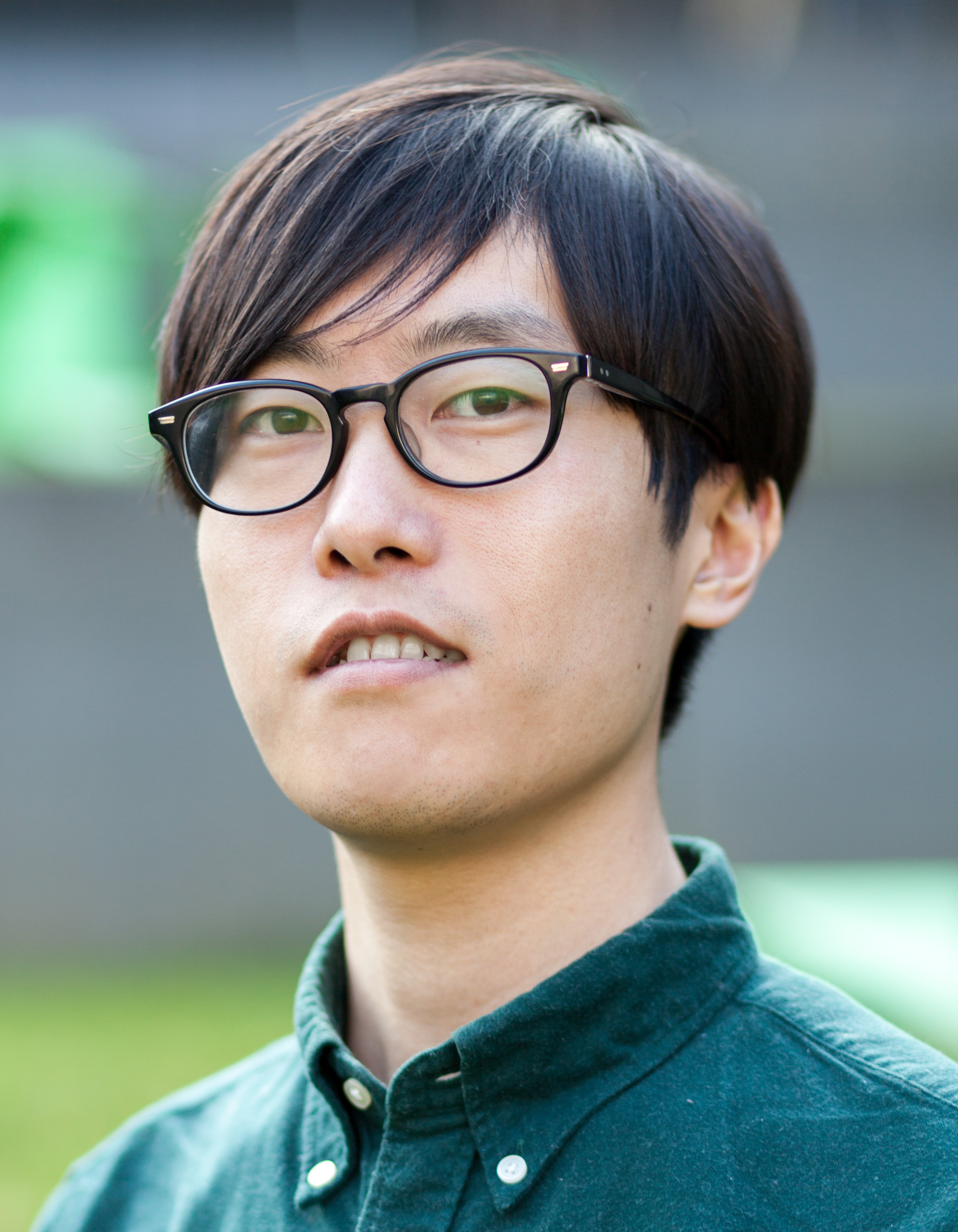}}]{Yusuke Sugano}
is Associate Professor at the Graduate School of Information Science and Technology, Osaka University, Japan.
Yusuke Sugano received his BSc., MSc., and PhD degrees from the University of Tokyo, in 2005, 2007, and 2010, respectively. 
He was previously a Postdoc in the Perceptual User Interfaces Group at the Max Planck Institute for Informatics, and a project research associate at Institute of Industrial Science, the University of Tokyo.
His research interests include computer vision and human-computer interaction.
\end{IEEEbiography}

\vspace{-1cm}

\begin{IEEEbiography}[{\includegraphics[width=1in,height=1.25in,clip,keepaspectratio]{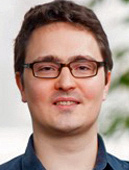}}]{Mario Fritz}
is Senior Researcher at the Max Planck Institute for Informatics, Germany, where he heads the Scalable Learning and Perception Group. Mario Fritz received his MSc. in Computer Science from the University of Erlangen-Nuremberg, Germany, in 2004 and his PhD from the Technical University of Darmstadt, Germany, in 2008. From 2008 to 2011, he was a Postdoc at the International Computer Science Institute (ICSI) and UC Berkeley, US funded by a Feodor Lynen Fellowship from the Alexander von Humboldt Foundation. His research interests include computer vision, machine learning, privacy, and natural language processing.
\end{IEEEbiography}

\vspace{-1cm}

\begin{IEEEbiography}[{\includegraphics[width=1in,height=1.25in,clip,keepaspectratio]{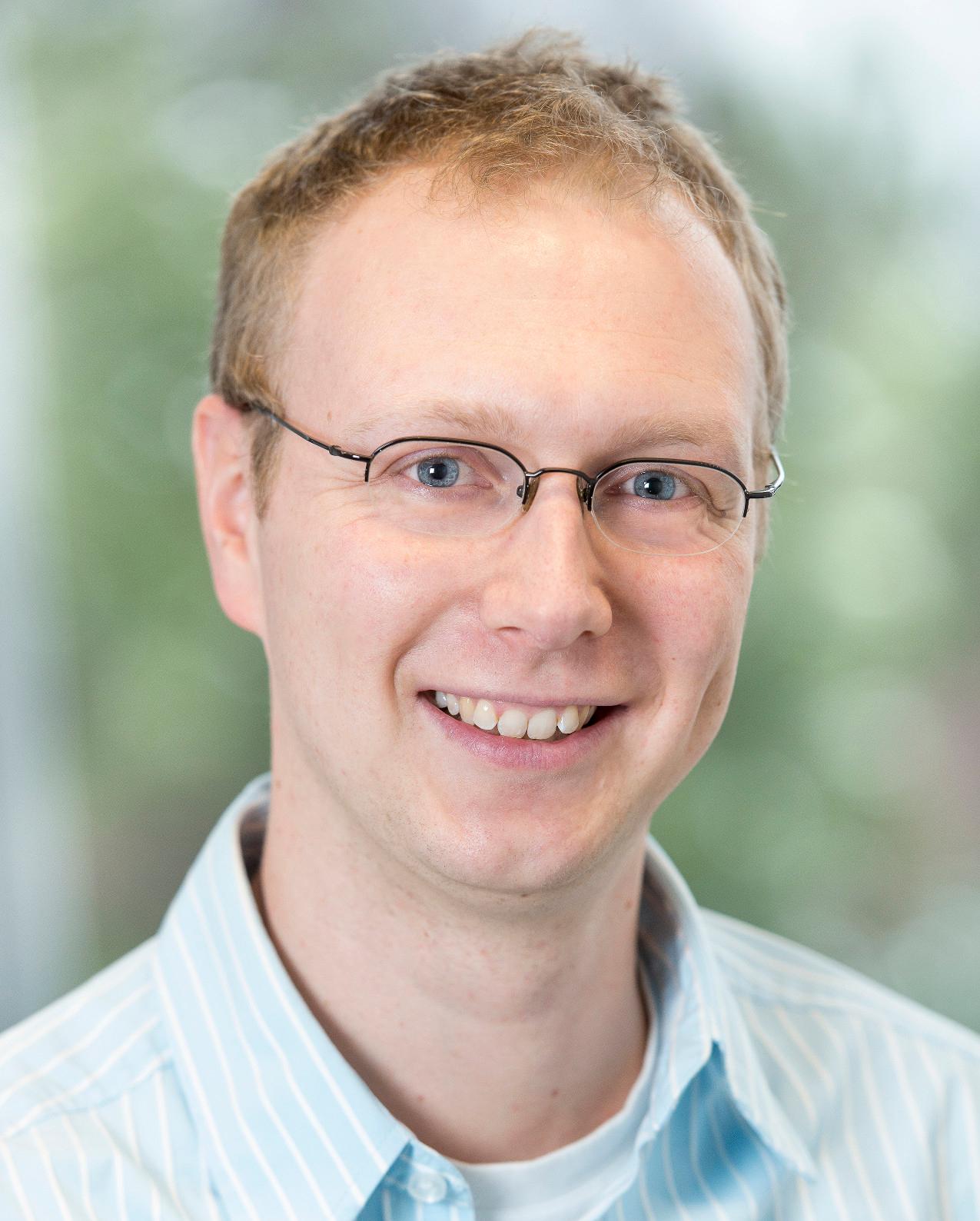}}]{Andreas Bulling}
is head of the Perceptual User Interfaces Group at the Max Planck Institute for Informatics and the Cluster of Excellence on Multimodal Computing and Interaction at Saarland University, Germany. Andreas Bulling received his MSc. in Computer Science from the Karlsruhe Institute of Technology, Germany, in 2006 and his PhD in Information Technology and Electrical Engineering from ETH Zurich, Switzerland, in 2010. From 2010 to 2013, he was a Feodor Lynen and Marie Curie Research Fellow at the University of Cambridge, UK.
His research interests include computer vision, ubiquitous computing, and human-computer interaction.
\end{IEEEbiography}

\end{document}